\documentclass{article}

\usepackage{graphicx}
\usepackage{amssymb}
\usepackage{latexsym}
\usepackage{graphicx}
\usepackage{epstopdf}
\usepackage{caption}
\usepackage{url}
\usepackage{xcolor}
\usepackage{amsmath}
\usepackage{color, colortbl}
\usepackage{enumitem}

 \usepackage{here}
\usepackage{lineno}
\usepackage[most]{tcolorbox}

\usepackage{PRIMEarxiv}

\usepackage[utf8]{inputenc} 
\usepackage[T1]{fontenc}    
\usepackage{hyperref}       
\usepackage{url}            
\usepackage{booktabs}       
\usepackage{amsfonts}       
\usepackage{nicefrac}       
\usepackage{microtype}      
\usepackage{lipsum}
\usepackage{fancyhdr}       
\usepackage{graphicx}       
\graphicspath{{media/}}     
\usepackage{makecell}
\usepackage{rotating}
\usepackage{array}

\title{Unlocking the Potential of ChatGPT: A Comprehensive Exploration of its Applications, Advantages, Limitations, and Future Directions in Natural Language Processing}

\author{
  Walid Hariri \\
  Labged Laboratory, Computer Science department\\ Badji Mokhtar University  \\
  Annaba, Algeria\\
  walid.hariri@univ-annaba.dz}

\begin{document}
\maketitle

\begin{abstract}
Large language models, pivotal in artificial intelligence, find diverse applications. ChatGPT (Chat Generative Pre-trained Transformer), an OpenAI creation, stands out as a widely adopted, powerful tool. It excels in chatbots, content generation, language translation, recommendations, and medical applications, due to its ability to generate human-like responses, comprehend natural language, and adapt contextually. Its versatility and accuracy make it a potent force in natural language processing (NLP). Despite successes, ChatGPT has limitations, including biased responses and potential reinforcement of harmful language patterns. This article offers a comprehensive overview of ChatGPT, detailing its applications, advantages, and limitations. It also describes the main advancements from GPT-3 to GPT-4 Omni, comparing them with other LLMs like LLaMA 3, Gemini and Deepseek. The paper underscores the ethical imperative when utilizing this robust tool in practical settings. Furthermore, it contributes to ongoing discussions on artificial intelligence's impact on vision and NLP domains, providing insights into prompt engineering techniques.
\end{abstract}


\section{Introduction}
\label{sec:intro}
Artificial intelligence (AI) has revolutionized the way we interact with machines and has transformed a wide range of industries. One of the most promising applications of AI is in the field of natural language processing (NLP), which involves the development of algorithms and models that can understand and generate human language. Among these NLP tools, ChatGPT (Generative Pre-trained Transformer) is a public tool developed by OpenAI that is based on the GPT language model
technology \cite{OpenAI}. It has emerged as a powerful and versatile tool for processing natural language.  

ChatGPT has been successfully applied in various real-world applications, making it a valuable asset in our daily lives. One example of its application is in the development of chatbots, which are used in customer service, technical support, and as virtual assistants \cite{haleem2023era}. These chatbots can interact with customers in a natural and human-like way, providing them with information and answering their questions. For example, the chatbot created by the National Health Service (NHS) in the UK uses ChatGPT to provide health advice and information to its users.

Another example of ChatGPT application is in content generation, such as news articles and product descriptions. By training the model on relevant data, it can generate high-quality, relevant content automatically. The Associated Press uses ChatGPT to generate financial news articles, and GPT-3 language model has been used to generate everything from poetry to computer code \cite{kashefi2023chatgpt}.

ChatGPT has also been used for language translation, personalized recommendations, and medical diagnosis and treatment. For instance, ChatGPT has been used to develop a model that can diagnose medical conditions or recommend appropriate treatments by analyzing large amounts of medical data \cite{munir2023artificial}.

However, despite its many applications, there are also limitations to ChatGPT. For example, it may produce biased responses and perpetuate harmful language patterns if not trained on diverse and inclusive data. Therefore, it is important to take into account ethical considerations when using ChatGPT in real-world applications.

In this article, we will explore the various applications of ChatGPT in real-life scenarios, its advantages, and its limitations. We will also discuss the ethical implications of using ChatGPT and ways to overcome potential limitations. Additionally, the article adds to the ongoing discussions about the effects of artificial intelligence on the domains of vision and NLP by offering valuable insights into prompt engineering techniques.

By understanding the potential of this powerful tool, we can make more informed decisions about how to use it effectively and responsibly in our daily lives.

\begin{figure}[h]
\centering
\includegraphics[width=0.90\textwidth]{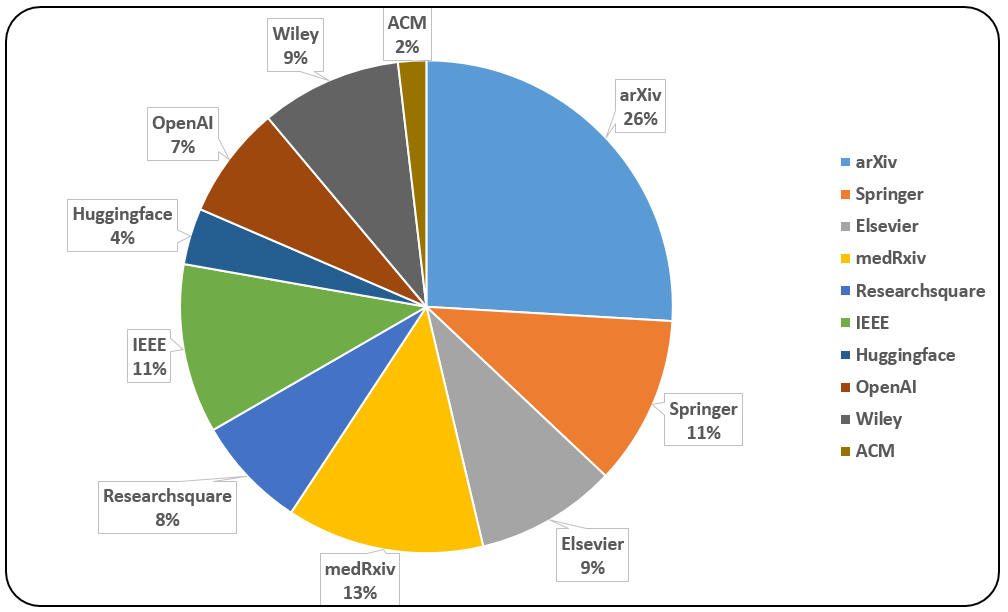}
\caption{Distribution of articles found in various publishers and pre-print repositories analyzed in this paper.}
\label{fig:sector}
\end{figure}

Various databases have been used in this review paper, including IEEE Xplore, ScienceDirect, SpringerLink, ACM Digital Library, and pre-print repositories such as Researchsquare, medRxiv, and ArXiv, to search for sources related to ChatGPT. They mainly focused on pre-print repositories and websites of OpenAI and Huggingface since the topic is still new and not yet widely published in international journals. They also included conference papers and used search engines like Google Scholar, Semantic Scholar, and Academia. Figure \ref{fig:sector} illustrates the percentage distribution of articles found in various publishers and pre-print repositories analyzed in this paper. The authors selected articles using specific keywords like ChatGPT, Advantages, Limitations, Large language models, NLP, Chatbots, OpenAI, Deep learning, pre-trained models, Text generation, Transformers, GPT-3.5, GPT-4, and ChatGPT applications. The most recent update of the paper using the aforementioned keywords was conducted on April 13, 2024.

The rest of the paper is organized as follows: Section \ref{sec:overview} presents an overview of ChatGPT and its capabilities. The transformer architecture and history are presented in Section \ref{trans}. In Section \ref{sec:app}, many applications of ChatGPT in real-word scenarios have been presented in detail with many examples. Section \ref{sec:ad} presents the advantages of ChatGPT in natural language processing, where the limitations are discussed in Section \ref{sec:dis}. Ethical considerations of ChatGPT are highlighted in Section \ref{sec:ethi}. The prompt engineering and generation are discussed in Section \ref{sec:prompt}. Sections \ref{sec:future1} and \ref{sec:future2} discuss the future directions of ChatGPT in NLP and vision domains. Conclusions end the paper.

\section{Overview of ChatGPT and its capabilities}
\label{sec:overview}
ChatGPT (Generative Pre-trained Transformer) is generative AI model (GAI) based on natural language processing techniques, developed by OpenAI, a leading AI research organization founded by a group of technology luminaries including Elon Musk, Sam Altman, Greg Brockman, and Ilya Sutskever \cite{openaichatgptdef}. This AI tool has achieved 100 million users in just three months. 
We differentiate between two main evolutions of ChatGPT based on the functionalities it offers, which are either focused solely on NLP or incorporate visual information. We also compare the recent advancement of ChatGPT with LLaMa 3 and Gemini.

\begin{figure}[h]
\centering
\includegraphics[width=0.99\textwidth]{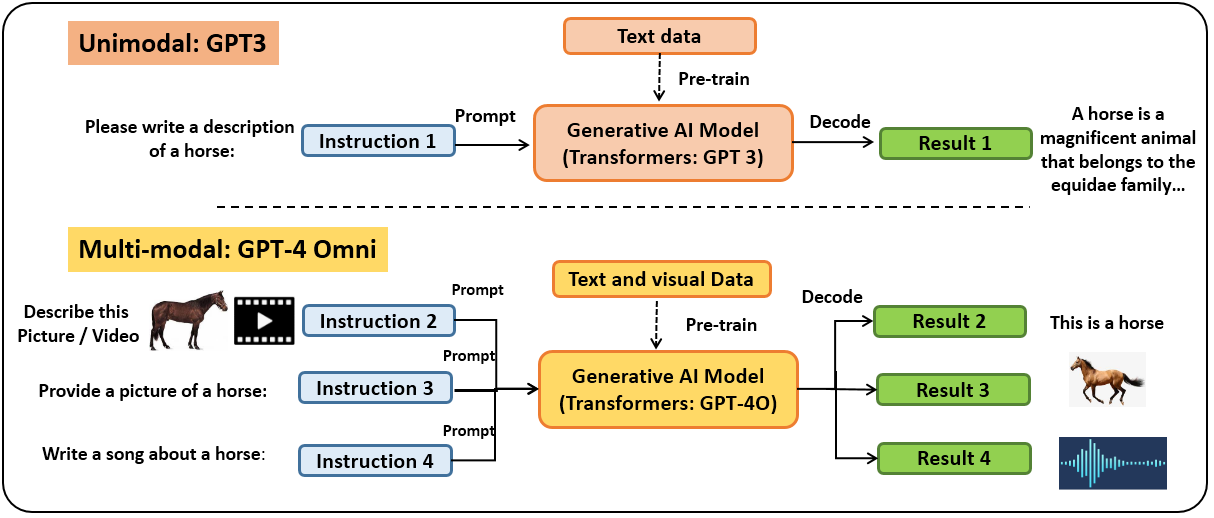}
\caption{Generative AI models: unimodal and multi-modal examples.}
\label{fig:gen}
\end{figure}

\paragraph{\textbf{NLP-based:}}
The development of ChatGPT was a major milestone in the field of NLP. Prior to its release, NLP models were typically task-specific and required significant amounts of labeled data for training. ChatGPT, on the other hand, was pre-trained on vast amounts of unlabeled data, allowing it to generate high-quality natural language text without specific task-related training data. 
The development of ChatGPT was based on the Transformer architecture, a neural network architecture designed specifically for NLP tasks. The Transformer architecture was introduced in a seminal paper by Vaswani et al. in 2017 \cite{vaswani2017attention} and quickly gained popularity due to its ability to outperform existing NLP models on a range of tasks.
In June 2018, OpenAI released the first version of ChatGPT, which was pre-trained on a massive dataset of over 40 GB of text data from the internet. The release of ChatGPT was met with significant excitement and attention from the NLP community, as it demonstrated the potential for pre-trained models to generate high-quality natural language text. There are two main categories of GAI models: unimodal models and multimodal models. Unimodal models take prompts from the same modality as the content they generate, while multi-modal models can accept prompts from different modalities and produce results in multiple modalities as shown in Figure \ref{fig:gen}.

Over the following years, OpenAI continued to develop and refine ChatGPT, releasing several larger and more advanced versions of the model. In May 2020, OpenAI released GPT-3, the latest and most powerful version of ChatGPT, which has been widely hailed as a breakthrough in NLP. GPT-3 has over 175 billion parameters, making it the largest NLP model to date, and has been used in a range of applications, from chatbots and virtual assistants to content generation and language translation. On March 14th, 2023, OpenAI announced the release of GPT-4 \cite{gpt4,katz2023gpt}, which has 100 trillion parameters that offer greater improvements compared to those found in GPT-3 such as accepting image and text inputs and emitting text outputs.

\begin{figure}[h]
\centering
\includegraphics[width=0.66\textwidth]{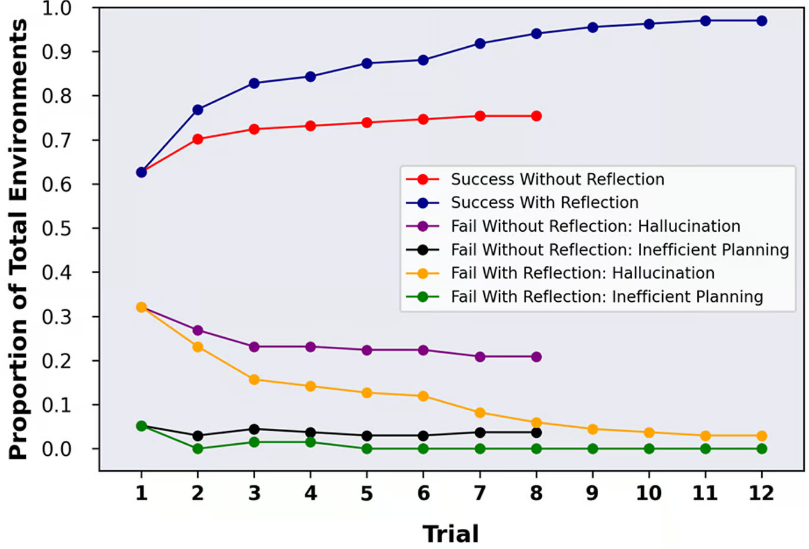}
\caption{GPT-4 performance boost using "Reflexion" technique \cite{reflexion}.}
\label{fig:reflexion}
\end{figure}

GPT-4 performs better than previous large language models and most current advanced systems on traditional NLP benchmarks. In addition, GPT-4 shows remarkable results on the MMLU benchmark, which includes multiple-choice questions covering 57 topics in English and other languages. GPT-4 not only surpasses existing models in English by a significant margin but also exhibits strong performance in other languages. On the translated versions of the MMLU benchmark, GPT-4 outperforms the state-of-the-art models presented in \cite{scao2022bloom}. Recently, it has been proven that GPT-4 becomes more accurate when it adopts self-supervising learning with 30\% of performance boost using the "Reflexion" technique. This means that the existing impressive capability of GPT-4 to carry out different tests is enhanced by a newly introduced framework that enables AI agents to imitate human-like self-reflection and self-evaluation \cite{reflexion,liu2023summary}. Essentially, it involves additional steps in which GPT-4 creates tests to analyze its own answers, identifying mistakes and weaknesses, and subsequently revises its solutions accordingly. As a result, this framework enhances the ability of GPT-4 to perform various tests more effectively as shown in Figure \ref{fig:reflexion}.

\paragraph{\textbf{NLP and visual-based:}}
GPT-4o ("o" stands for "omni") introduced by Open AI on May 14, 2024, represents a significant advancement in creating more natural human-computer interactions. It can take input in various forms—text, audio, images, and video—and generate outputs in text, audio, or images \cite{GPT4o}. Impressively, it can respond to audio inputs in as little as 232 milliseconds, with an average response time of 320 milliseconds, which is comparable to human response times in conversation. In terms of performance, GPT-4o matches GPT-4 Turbo for text processing in English language teaching \cite{pang2024chatgpt} and coding tasks, while offering enhanced capabilities in non-English text processing. Additionally, it is notably faster and 50\% more cost-effective in the API. GPT-4o also excels in understanding vision and audio compared to previous models. Below, we highlight the key enhancements of GPT-4o compared to previous versions.

\begin{enumerate}
    \item \textbf{Multimodal Integration}: GPT-4o can process and generate text, images, and audio, allowing for richer and more interactive responses. This makes it versatile for various applications, including education, customer support, and content creation \cite{uniteai,datacamp}.
    
    \item \textbf{Improved Contextual Awareness}: It maintains context over longer conversations, providing more coherent and contextually relevant responses. This enhancement is particularly beneficial for personal assistants and customer support systems \cite{uniteai}.
    
    \item \textbf{Enhanced Language Understanding}: Trained on a larger and more diverse dataset, GPT-4o offers improved accuracy and coherence in text generation. This makes it more effective in handling complex conversations and nuanced topics \cite{uniteai}.
    
    \item \textbf{Advanced Tokenization}: GPT-4o features a better tokenization model, especially for non-Roman alphabets, which reduces the number of tokens needed and speeds up text generation. This also makes the model more cost-efficient \cite{datacamp}.
    
    \item \textbf{Cost-Efficiency}: The model is optimized to be 50\% more cost-efficient than its predecessors, making sophisticated AI tools more accessible to a broader range of users \cite{geeky-gadgets}.
    
    \item \textbf{Real-Time Collaboration}: GPT-4o supports real-time collaboration features, allowing multiple users to interact with the AI simultaneously, which is useful in team environments and project management \cite{uniteai}.
    
    \item \textbf{Customization and Fine-Tuning}: Users can tailor the model to better suit specific applications by adjusting its behavior, tone, and response style, ensuring that it aligns more closely with unique requirements \cite{uniteai}.
    
    \item \textbf{Desktop Application}: Alongside the launch of GPT-4o, OpenAI introduced a ChatGPT desktop app, enhancing usability and integration into various workflows \cite{datacamp}.
\end{enumerate}

From the above enhancements compared to previous versions, GPT-4o can impressively analyze and interpret visual information from images and videos, enabling it to provide detailed descriptions, recognize objects, and even understand complex scenes. This multimodal capability enhances its usability in various applications such as security, content moderation, and educational tools. Visual capabilities are described in the following.
    \begin{itemize}
        \item \textbf{Image Analysis}: When provided with a photo, GPT-4o can describe the contents of the image, identify objects, and even detect emotions from facial expressions. This makes it useful for applications like social media content moderation, automated tagging, and accessibility tools for visually impaired users.
        \item \textbf{Video Analysis}: GPT-4o can process video inputs from cameras, offering functionalities like real-time object tracking, action recognition, and scene understanding. This enhancement is valuable in security systems, where the AI can monitor video feeds for unusual activities or potential threats, and in automated video editing tools that can identify key moments in footage.
        \item \textbf{Educational Applications}: In educational settings, GPT-4o's ability to incorporate visual aids into its responses can significantly enhance learning. For instance, when explaining a scientific concept, it can include relevant diagrams or video snippets, making complex ideas easier to grasp.
        \item \textbf{Healthcare}: In healthcare, GPT-4o's visual analysis capabilities can assist in diagnosing medical images, such as X-rays or MRIs, by identifying anomalies and providing preliminary assessments, thereby aiding medical professionals.
        \item \textbf{Custom Visual Model Training}: Users can fine-tune GPT-4o's visual processing abilities to suit specific needs. For instance, a business could train the model to recognize their products in images or videos, streamlining inventory management and quality control processes.
    \end{itemize}

The introduction of GPT-4o marks a pivotal moment in AI development, combining increased accessibility, improved performance, and enhanced multimodal capabilities to push the boundaries of human-machine interaction \cite{uniteai,geeky-gadgets}.

\paragraph{\textbf{Comparison with LLaMA 3:}}

LLaMA 3, the third iteration of the LLaMA (Large Language Model Meta AI) series, was introduced by Meta (formerly Facebook) in 2024 \cite{dubey2024llama}. It builds on the advancements of its predecessors, LLaMA 1 and LLaMA 2, incorporating improvements in model architecture, training techniques, and scalability. LLaMA 3 is designed to enhance performance across a variety of natural language processing tasks and to address some of the limitations observed in earlier models with 70B parameter scale.
While GPT-3 improves upon GPT-2 primarily by increasing the amount of training data, LLaMA-3 follows a more extensive development trajectory. Unlike the incremental improvements seen from GPT to GPT-2 to GPT-3, LLaMA-3 represents a broader evolution in model design, incorporating more complex and capable configurations \cite{kumar2024towards}. The development of LLaMA-3 involves multiple stages, including development, training, testing, and fine-tuning across various tasks. These tasks utilize a combination of supervised learning with and without reward functions, transfer learning, and reinforcement learning (RL) techniques \cite{xu2024magpie}. The model also introduces incremental modifications to the GPT architecture, enhancing its scalability and performance.

OpenAI's original GPT, released in 2016, was designed for supervised and semi-supervised training using large datasets, with fine-tuning achievable with minimal examples. It employed a transformer architecture based on attention mechanisms, which proved effective for processing human language and complex tasks. In contrast, LLaMA-3 utilizes a different transformer-based model known as the Linear Modular Unit (LMU), which was initially designed for control systems but later adapted for human language processing \cite{kalyan2023survey,haque2022brief,black2022gpt}. The LMU combines the strengths of successful language models while offering scale-invariance similar to Long Short-Term Memory (LSTM) networks, making it highly efficient for handling increasingly complex language tasks. Table \ref{tab:benchmark} provides a comparative analysis of several LLMs across various benchmarks and shot configurations. It includes five models: GPT-3.5, GPT-4, PaLM, PaLM-2-L, and LLaMA 2. This table highlights that GPT-4 generally performs better across most benchmarks, while models like PaLM-2-L and LLaMA 2 show strong performances in specific tasks.

\begin{table}[h!]
\centering
\caption{Benchmark comparison for various LLM models.}
\label{tab:benchmark}
\begin{tabular}{|l|c|c|c|c|c|}
\hline
\textbf{Benchmark (shots)} & \textbf{GPT-3.5} & \textbf{GPT-4} & \textbf{PaLM} & \textbf{PaLM-2-L} & \textbf{LLaMA 2} \\ \hline
MMLU (5-shot) & 70.0 & 86.4 & 69.3 & 78.3 & 68.9 \\ \hline
TriviaQA (1-shot) & - & - & 81.4 & 86.1 & 85.0 \\ \hline
Natural Questions (1-shot) & - & - & 29.3 & 37.5 & 33.0 \\ \hline
GSM8K (8-shot) & 57.1 & 92.0 & 56.5 & 80.7 & 56.8 \\ \hline
HumanEval (0-shot) & 48.1 & 67.0 & 26.2 & - & 29.9 \\ \hline
BIG-Bench Hard (3-shot) & - & - & 52.3 & 65.7 & 51.2 \\ \hline
\end{tabular}

\end{table}

\paragraph{\textbf{Comparison with Gemini:}}

The Gemini model, developed by Google DeepMind, represents a significant advancement in the field of large language models. Four distinct versions of Gemini have been identified: Ultra, Pro, Flash, and Nano introduced on December 6, 2023, each tailored to different performance and application needs \cite{Gemini}. Originally known as "Gemini" during its development phase, it was officially released as Gemini 1, reflecting Google's ongoing innovation in artificial intelligence. The Gemini model is part of Google's strategic efforts to maintain a leading position in AI development, competing directly with other top LLMs such as OpenAI's GPT series and Meta's LLaMA models.

Gemini is built on a sophisticated architecture designed to enhance its ability to understand and generate human-like text across a wide range of natural language processing tasks, including text generation, summarization, machine translation, and conversational AI \cite{guven2024performance}. The model emphasizes advanced reasoning capabilities, allowing it to tackle complex tasks that require a deep understanding of context and nuance. One of Gemini's key features is its focus on improving response accuracy and reducing biases, a common challenge in earlier LLMs \cite{islam2024gemini}. It integrates state-of-the-art techniques in AI ethics and safety, aiming to minimize harmful outputs while maintaining high performance in generating relevant and contextually appropriate responses. With its scalable and efficient architecture, Gemini 1 is optimized for processing large datasets and delivering quick, accurate responses, leveraging Google's vast computational resources. As part of a broader initiative, Gemini is deeply embedded within Google’s ecosystem, enhancing services like Google Search, cloud platforms, and other tools where advanced natural language understanding is critical. The release of Gemini marks a pivotal moment in Google's AI strategy, ensuring competitiveness in a rapidly advancing industry. Unlike standalone models, Gemini is expected to evolve as a foundation for future iterations, pushing the boundaries of AI’s ability to understand and interact with human language.
When compared to ChatGPT and LLaMA 3, Gemini distinguishes itself with its advanced reasoning and deep integration within Google's ecosystem. Table \ref{tab:LLaMA_Gemini} presents a comparison between LLaMA 3, Gemini Pro and Claude 3 Sonnet models across five benchmarks: MMLU, GPQA, HumanEval, GSM-8K, and MATH. Meta Llama 3 70B generally scores higher in MMLU, HumanEval, and GSM-8K, while Gemini Pro 1.5 leads in GPQA and MATH.

While ChatGPT is widely recognized for its versatility, ease of integration, and proven reliability across diverse applications, Gemini offers a unique focus on ethical AI practices, bias reduction, and seamless functionality within Google’s expansive range of services. Both models represent the cutting edge of AI technology, yet they cater to slightly different needs within the AI landscape.

\paragraph{\textbf{Comparison with Deepseek:}}
The advent of DeepSeek marks a significant milestone in the evolution of artificial intelligence, particularly in the realm of natural language processing \cite{guo2024deepseek,liu2024deepseek}. Developed by a team of researchers from China, the start-up DeepSeek was officially launched in early 2023, where the model DeepSeek-R1 was introduced in 2025. The project was spearheaded by a group of AI experts from leading Chinese universities and tech companies, who aimed to create a language model that could rival existing models like ChatGPT while being more cost-effective and accessible. DeepSeek is backed by a startup that focuses on leveraging cutting-edge AI technologies to provide scalable and affordable solutions for both businesses and individual users.

One of the key advantages of DeepSeek is its cost efficiency \cite{lu2024deepseek}. Unlike many other high-performance language models that require substantial computational resources and financial investment, DeepSeek has been optimized to deliver comparable performance at a fraction of the cost. This makes it an attractive option for startups, researchers, and organizations with limited budgets, enabling broader adoption of advanced NLP technologies.

While both DeepSeek and ChatGPT are built on advanced transformer architectures, they exhibit distinct characteristics in terms of performance, application, and user experience. Below, we outline some key points of comparison:

\begin{itemize}
    \item \textbf{Training data and scope:} 
    ChatGPT is trained on a diverse dataset that includes a wide range of internet text, enabling it to handle a variety of topics with reasonable proficiency. DeepSeek, on the other hand, leverages a more specialized and curated dataset, which allows it to excel in specific domains, particularly those requiring deep technical or scientific knowledge.
    
    \item \textbf{Response quality:} 
    ChatGPT is known for its ability to generate coherent and contextually relevant responses, making it suitable for general-purpose applications. DeepSeek, however, often produces more nuanced and detailed answers, particularly in complex scenarios where depth of understanding is crucial.
    
    \item \textbf{User interaction:} 
    ChatGPT has been widely praised for its user-friendly interface and ease of interaction, making it accessible to a broad audience. DeepSeek, while equally intuitive, offers additional customization options that cater to users with more specialized needs, such as researchers and professionals in technical fields.
    
    \item \textbf{Adaptability:} 
    Both models are capable of fine-tuning and adaptation to specific tasks. However, DeepSeek's architecture is designed to be more flexible, allowing for quicker and more efficient adjustments to new domains or tasks, which can be a significant advantage in rapidly evolving fields.
    
    \item \textbf{Cost efficiency:} 
    One of the standout features of DeepSeek is its cost-effectiveness \cite{krause2025deepseek,wu2024deepseek}. The model has been optimized to reduce computational overhead, making it less expensive to deploy and maintain compared to ChatGPT. This cost advantage is particularly beneficial for smaller organizations and individual researchers who may not have the resources to invest in more expensive AI solutions.
\end{itemize}

While ChatGPT remains a versatile and widely-used tool for general NLP tasks, DeepSeek introduces a new level of sophistication and specialization, particularly in areas requiring deep expertise. The choice between the two models ultimately depends on the specific requirements of the task at hand, with each offering unique strengths that cater to different user needs.

\begin{table}[h!]
\centering
\small
\caption{Performance comparison of Meta Llama 3 70B, Gemini Pro 1.5, and Claude 3 Sonnet}
\label{tab:LLaMA_Gemini}
\begin{tabular}{|l|c|c|c|}
\hline
 & \textbf{Meta LLama 3 70B} & \textbf{Gemini Pro 1.5} & \textbf{Claude 3 Sonnet} \\
\hline
\textbf{MMLU (5-shot)} & 82.0 & 81.9 & 79.0 \\
\hline
\textbf{GPQA (0-shot)} & 39.5 & 41.5 & 38.5 \\
\hline
\textbf{HumanEval (0-shot)} & 81.7 & 71.9 & 73.0 \\
\hline
\textbf{GSM-8K (8-shot, CoT)} & 93.0 & 91.7 & 92.3 \\
\hline
\textbf{MATH (4-shot, CoT)} & 50.4 & 58.5 (Minerva prompt) & 40.5 \\
\hline
\end{tabular}

\end{table}

It is worth noting that GPT-3 and data-to-text are both technologies that belong to the field of NLG or "Natural Language Generation". This term refers to the process of generating text in natural language through automation. Although these two technologies may appear similar at first, they operate in distinct ways. Table \ref{tab:diff} presents some of the differences between them. 

\begin{table*}[h!]
\footnotesize
\caption{GPT-3 versus data-to-text.}
\begin{center}
\label{tab:diff}
\begin{tabular}{|*{10}{c|}}
\hline
\textbf {Method}           & \textbf {Data-to-Text	}     & \textbf {GPT-3}  \\ \hline
\textbf{Text Generation}  & \makecell{Generates text based on structured data \\ such as attributes from tables like product \\features or soccer match results}	 & \makecell{Learns from existing text through training\\ on hundreds of billions of words from sources \\like Wikipedia, books, and web pages}
 \\ \hline
\textbf{\makecell{Control Over\\ Content}}  &  User has full control over the resulting text	        & \makecell{Generated content cannot \\be controlled by the user}
\\ \hline
\textbf{Text Quality} & \makecell{Emphasis on consistency, meaningfulness,\\ and overall quality}	    & \makecell{Texts must be fact-checked as they may\\ contain incorrect or inappropriate \\information}
\\  \hline
\textbf{Scalability} &  Texts are easily scalable and customizable	   & \makecell{Generates individual texts, but not as\\scalable as Data-to-Text}
\\  \hline
\textbf{Languages} &  \makecell{Can create multilingual content in up to 110 \\languages}	   & \makecell{Can create multilingual content on a \\per-language basis}
\\  \hline
\textbf{Usage} &  \makecell{Best for generating large amounts of text from \\ structured data sets with variable details}	   &\makecell{Useful for creating basic text and \\simplifying the writing process.}

 \\  \hline
		\end{tabular}
\end{center}
\end{table*}

\section{Transformers and pre-trained language models}
\label{trans}
\begin{figure}[h]
\centering
\includegraphics[width=0.43\textwidth]{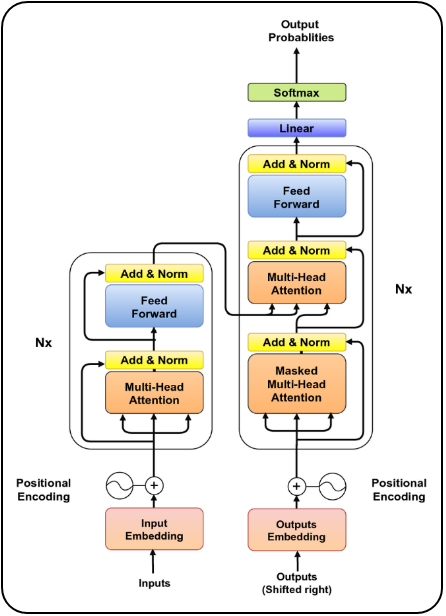}
\caption{Transformer architecture.}
\label{fig:trans}
\end{figure}

The Transformer architecture is used as the main structure for several cutting-edge models, including GPT-3 \cite{brown2020language}, DALL-E-2 \cite{ramesh2022hierarchical} and Codex \cite{chen2021evaluating}. It was created to overcome the limitations of traditional models like RNNs in managing sequences of varying lengths and contextual information. The Transformer relies on a self-attention mechanism, which enables the model to focus on different segments of the input sequence (See Figure \ref{fig:trans}). The Transformer comprises an encoder and a decoder. The encoder processes the input sequence and produces hidden representations, while the decoder uses these hidden representations to generate the output sequence. Each layer of the encoder and decoder contains a multi-head attention mechanism and a feed-forward neural network. The multi-head attention is the most important part of the Transformer, as it determines how tokens are weighted based on their relevance. This method of information routing allows the model to better handle long-term dependencies, leading to improved performance across various NLP tasks. Another advantage of the Transformer is its parallelizability, which allows it to handle large-scale pre-training and adaptability to different downstream tasks without inductive biases. Google recently unveiled Gemma, a family of lightweight open models designed as a competitor to ChatGPT. Gemma serves as the foundation for the Gemini models, which are based on a transformer decoder architecture. It offers two sizes of models with either 2 billion or 7 billion parameters \cite{team2024gemma}. The release includes both pretrained and fine-tuned checkpoints for users. Multi-Query Attention \cite{shazeer2019fast}, RoPE Embeddings \cite{su2024roformer}, GeGLU Activations \cite{shazeer2020glu}, and RMSNorm \cite{zhang2019root} are some improvements used to build the Gemma model. In April 2024, RecurrentGemma \cite{RecurrentGemma24} that uses the novel Griffin architecture of Google is introduced with fewer trained tokens than Gemma-B2. 
GPT-Neo, developed by the open-source community EleutherAI, is an accessible alternative to GPT-3. Founded in 2020, EleutherAI aims to democratize AI by creating and sharing open-source models with 2.7 B parameters \cite{GPT-Neo}. GPT-J \cite{GPT-J} is an open-source large language model created by EleutherAI, designed to offer high performance similar to GPT-3. It has 6 billion parameters and is known for its ability to generate coherent and contextually relevant text. It is freely available, supporting research and development in AI.

\begin{figure}[h]
\centering
\includegraphics[width=0.95\textwidth]{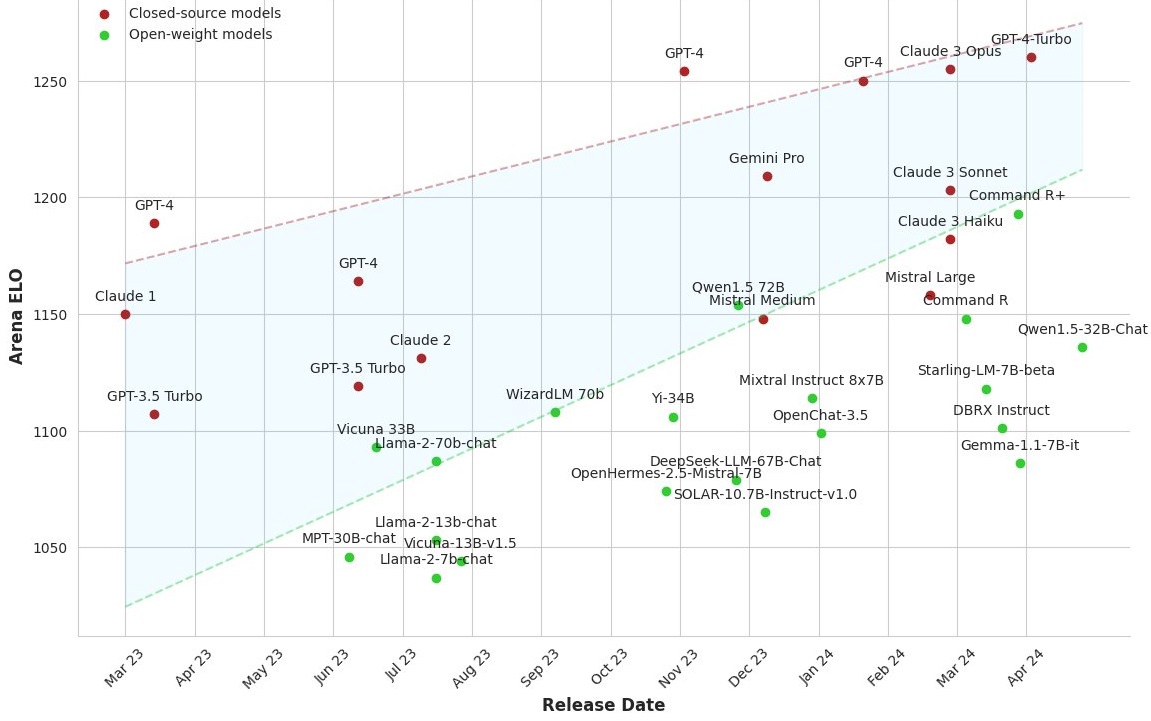}
\caption{Open-source and closed-source models according to Arena Elo rating system.}
\label{fig:close_openfig}
\end{figure}

Open-source models thrive on community collaboration, fostering rapid innovation and broader applicability through diverse contributions from developers, researchers, and enthusiasts. This collaborative environment accelerates the refinement and expansion of these models to address a wide range of problems. In contrast, closed-source models benefit from structured, corporate-backed development, which provides the advantage of faster deployment, comprehensive support, and targeted solutions tailored to specific business needs. Despite these benefits, the debate over which approach is superior persists, as each has its strengths and limitations depending on the context and objectives. Table \ref{tab:comparison_open_closed} compares open-source and closed-source models across various aspects, including development model, contributors, innovation, and provides examples for each category. Figure \ref{fig:close_openfig} presents a comparison between the best-performing closed-source and open-source models reported by Arena Elo rating system between Mar 2023 and Apr 2024.

\begin{table}[h!]
\centering
\footnotesize
\caption{Comparison of open source and closed source models}
\label{tab:comparison_open_closed}
\begin{tabular}{|p{3cm}|p{6cm}|p{6cm}|}
\hline
\textbf{Aspect} & \textbf{Open source models} & \textbf{Closed source models} \\
\hline
\textbf{Development Model} & Community-driven, collaborative, iterative & Centralized, structured, product-focused \\
\hline
\textbf{Contributors} & Developers, researchers, and enthusiasts & Corporate teams with specialized roles \\
\hline
\textbf{Innovation} & Rapid iteration, diverse perspectives, and unforeseen evolution & Structured, focused on specific business needs \\
\hline
\textbf{Applicability} & Versatile, adaptable to a wide range of problems & Tailored to specific business applications and market needs \\
\hline
\textbf{Ownership} & Shared among contributors, fostering investment and engagement & Owned by a corporate entity with clear control over the development \\
\hline
\textbf{Resource access} & Community-driven, often limited resources & Backed by extensive corporate resources and technical expertise \\
\hline
\textbf{Documentation and support} & Variable, depending on community contributions & Comprehensive, usually includes training and support \\
\hline
\textbf{Deployment speed} & Can be slower due to the need for community-driven refinement & Typically faster due to structured development and corporate backing \\
\hline
\textbf{Market readiness} & Requires additional effort to commercialize & Generally product-validated and market-ready \\
\hline
\textbf{Efficacy and customization} & High potential for efficacy due to diverse contributions and use cases & High efficiency for specific business needs and integration \\
\hline
\textbf{Customer understanding} & Deep understanding from contributors who may also use the models & Often based on the needs of the corporation and its clients \\
\hline
\textbf{Examples} & GPT-Neo, GPT-J, LLaMA, Bloom, T5 & GPT-4, Bard, Claude (Anthropic), Gemini \\
\hline
\end{tabular}
\end{table}

The Transformer architecture has become the primary choice in natural language processing due to its ability to learn and parallelize. Pre-trained language models that use the Transformer architecture can be divided into two categories based on their training tasks: autoregressive language modeling and masked language modeling.

\textbf{Autoregressive language modeling:} used in models like GPT-3 and Open pre-trained transformer language models (OPT) \cite{zhang2022opt}, involves modeling the probability of the next token in a sentence given the preceding tokens, making it a left-to-right language modeling approach. Autoregressive models are better suited for generative tasks than masked language models. RoBERTa uses the same architecture as BERT but performs better by increasing the amount of pre-training data and incorporating more challenging pre-training objectives. XL-Net \cite{yang2019xlnet}, another model based on BERT, uses permutation operations to change the prediction order for each training iteration, allowing the model to learn more information across tokens. 

\begin{figure}[h]
\centering
\includegraphics[width=0.96\textwidth]{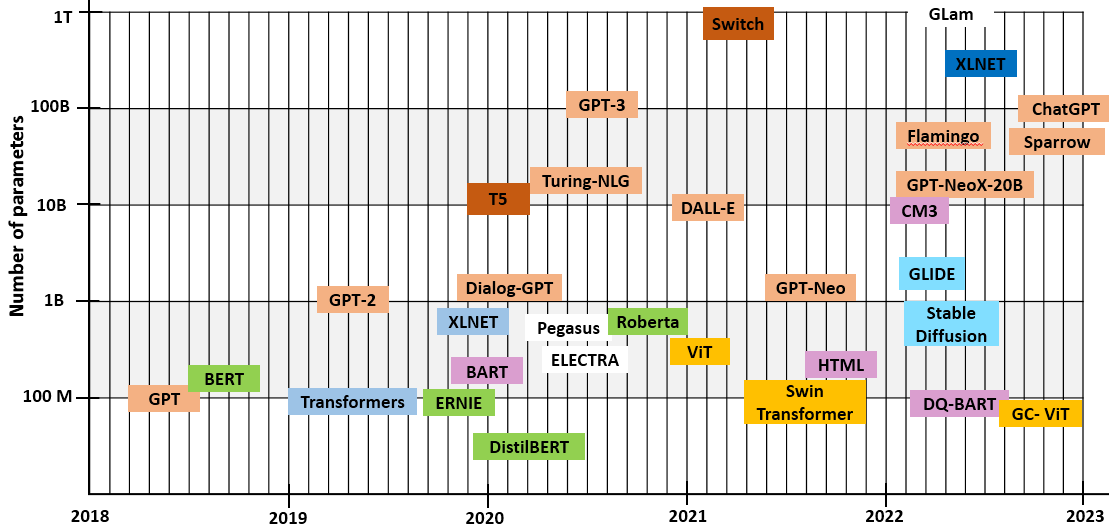}
\caption{Transformer timeline.}
\label{fig:time}
\end{figure}

\textbf{Masked language modeling:} used in models like BERT \cite{devlin2018bert} and RoBERTa \cite{liu2019roberta}, involves predicting the probability of a masked token given its context within a sentence. Figure \ref{fig:time} depicts the timeline of the most known generative models with their number of parameters.

\section{Applications of ChatGPT in real-world scenarios}
\label{sec:app}
Since its emergence, ChatGPT has become increasingly popular and has been applied in a wide range of applications and fields. Some notable examples include healthcare and education chatbots, finance, entertainment, cybersecurity, marketing, and vision tasks. While the main modality used in ChatGPT is text, it is also possible to incorporate other tools and create a multimodal application that includes sound, images, or videos. Some examples of ChatGPT use cases are presented in Figure \ref{fig:cases}, which demonstrate how ChatGPT can be applied in real-world scenarios, explained one by one in the following:

\begin{figure}[h]
\centering
\tcbox[colframe=gray!30!black]{\includegraphics[width=0.53\textwidth]{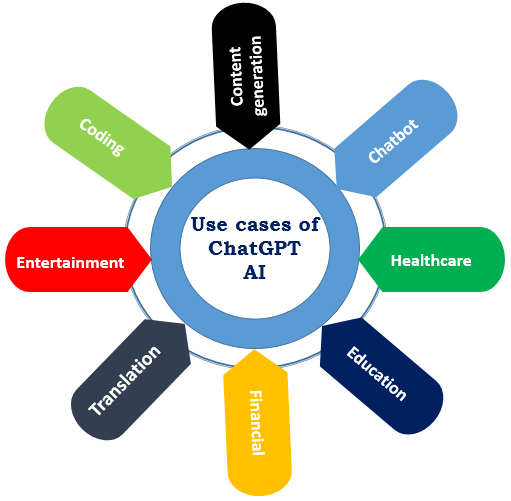}}
\caption{Use cases of ChatGPT.}
\label{fig:cases}
\end{figure}

\subsection{Healthcare:} ChatGPT has been used to develop virtual health assistants that can understand and respond to patients' questions and concerns \cite{biswas2023role}. The following are some instances of how artificial intelligence is utilized in the healthcare industry as described in \cite{munir2023artificial}:

\begin{itemize}
    \item \textbf{Diagnostic support:} AI algorithms can aid healthcare providers in diagnosing various medical conditions, such as skin cancer, heart disease, and eye diseases.
   \item \textbf{Predictive analytics:} By analyzing patient data, AI can predict future health issues and help healthcare providers to intervene early.
    \item \textbf{Personalized medicine:} AI can analyze an individual patient's data to create customized treatment plans to improve patient outcomes.
    \item \textbf{Imaging analysis:} AI algorithms can assist in the interpretation of medical images like X-rays, CT scans, and MRI scans.
    \item \textbf{Drug discovery:} AI can analyze large amounts of data from clinical trials to expedite drug development and enhance the chances of success for new treatments.
    \item \textbf{Telemedicine:} AI supports remote patient consultations and improve healthcare accessibility in remote areas.
    \item \textbf{Surgical support:} AI can help during surgical procedures by guiding surgical instruments and providing real-time feedback.
\end{itemize}

\begin{figure}[h]
\centering
\tcbox[colframe=gray!30!black,
           colback=gray!20]{\includegraphics[width=0.55\textwidth]{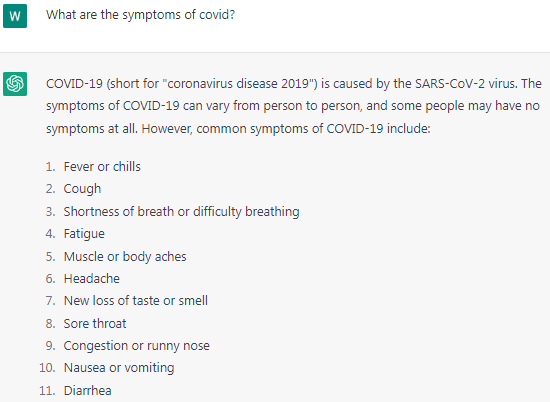}}
\caption{Example of medical question about Covid-19 symptoms.}
\label{fig:covid}
\end{figure}

ChatGPT can significantly enhance the quality of existing services by offering more accurate and tailored assistance to both users and service providers. By leveraging advanced natural language processing capabilities, ChatGPT can understand and respond to complex queries with precision, thereby improving the efficiency and effectiveness of customer support, virtual assistants, and other interactive services. For example, Figure \ref{fig:covid} shows the answer of ChatGPT about Covid-19 symptoms. These assistants can provide personalized recommendations and advice, such as medication reminders or diet plans, and help patients manage their health conditions. Moreover, ChatGPT has the potential to generate discharge summaries, which are comprehensive medical documents summarizing a patient's hospital stay. More Potential Benefits of ChatGPT in Healthcare can be found in \cite{health}. Wang et al. \cite{wang2023chatgpt} proposed a study to compare the level of knowledge and interpretation skills between ChatGPT and medical students in China. To achieve this goal, the Chinese National Medical Licensing Examination was administered to both ChatGPT and medical students, and their performances were compared. Johnson et al. \cite{johnson2023assessing} assessed the accuracy and reliability of ChatGPT to generate medical responses. In this study, a total of 33 physicians from 17 different specialties were involved in the generation of 284 medical questions. These questions were then categorized by the physicians themselves based on their perceived difficulty level, which was classified as either easy, medium, or hard. The questions were designed to elicit either binary (yes/no) responses or descriptive answers.

On the other hand, Benoit \cite{benoit2023chatgpt} has tested ChatGPT to assess its potential in quickly generating, rewriting, and evaluating sets of clinical vignettes. The evaluation included diagnostic and triage accuracy of many diseases including anaphylaxis, asthma, bronchiolitis, croup, ear infection, fever, functional
constipation, and gastroenteritis. Similarly, Gunther Eysenbach\cite{eysenbach2023role} conducted experiments using ChatGPT to explore possible applications of chatbots in medical education. ChatGPT demonstrated its ability to create a virtual patient simulation and quizzes for medical students, evaluate a simulated doctor-patient interaction, and even summarize a research article.
Usually, junior doctors are responsible for composing these summaries, but due to their workload, they may not be given the priority they deserve, causing delays in patients' discharges and incomplete summaries \cite{dahmen2023artificial}. This puts pressure on an already overworked junior doctor workforce and may lead to potential patient safety issues during the transition from secondary to primary care. However, if ChatGPT is utilized, it could alleviate the burden of composing discharge summaries, resulting in more timely and high-quality summaries that are essential for safe patient care \cite{patel2023chatgpt}. Other ChatGPT decision-making and clinical decision support can be found in \cite{ali2023performance,rao2023evaluating,liu2023assessing,rao2023assessing,chen2023utility,strong2023performance,raile2024usefulness,ferdush2023chatgpt}.

\subsection{Education:} ChatGPT has been used to develop intelligent tutoring systems that can provide personalized learning experiences to students \cite{cai2024factors}. These systems can understand students' learning styles and adapt the content and teaching methods to their needs, helping them to achieve better learning outcomes \cite{zhai2022chatgpt}. 
 To provide further understanding of the impact of ChatGPT on students, Raman et al. \cite{raman2023university} conducted a study involving 288 university students, aims to identify the factors that determine students' intentions to use ChatGPT in higher education, using Rogers' perceived theory of attributes as a theoretical framework. The study examined five factors that influence the adoption of ChatGPT: Relative Advantage, Compatibility, Ease of Use, Observability, and Trialability. The gender-based analysis of the study indicates that male students prioritize compatibility, ease of use, and observability as factors in ChatGPT adoption. On the other hand, female students give more importance to factors such as ease of use, compatibility, relative advantage, and trialability when it comes to adopting ChatGPT.

AlAfnan et al. \cite{alafnan2023chatgpt} investigate in their research the potential advantages and difficulties of utilizing the ChatGPT chatbot for academic purposes. The authors aim to provide recommendations to teachers and professors in schools and universities. The study aims to answer three main questions: (1) What are the potential benefits of using ChatGPT for academic purposes? (2) What are the challenges associated with using ChatGPT for academic purposes? and (3) What advice can be given to instructors who use ChatGPT in their teaching?. Further, in this study, the authors tested the similarity index of generated text by ChatGPT and paraphrased sentences using Turnitin. Additionally, Cooper \cite{cooper2023examining} has analyzed the use of ChatGPT in science education by addressing three questions: (1) How does ChatGPT respond to inquiries regarding science education? (2) What are some potential ways educators can integrate ChatGPT into their science teaching? and (3) How was ChatGPT utilized in this research, and what are the author's reflections on its use as a research tool?

Moreover, The impact of ChatGPT on education has raised concerns, as it is capable of writing essays on various topics. To test its abilities, Is has been assigned an exam and a final project from a class on science denial at George Washington University. Although it was able to find factual answers, its scholarly writing skills still need improvement. This could prompt educators to rethink their teaching methods and assignments to encourage creativity and critical thinking, rather than relying on AI \cite{tlili2023if}. While this could be positive, there are concerns about the use of ChatGPT in scientific paper writing. A recent study found that academic reviewers only detected 63\% of the fake abstracts generated by ChatGPT. This raises the possibility of AI-generated text being published in scientific literature, which is a worrisome trend \cite{thorp2023chatgpt2}. Besides, the study showed that the ability of ChatGPT to provide specific and relevant information on various topics such as science, history, business, health, and technology was perceived as useful by many users. One participant even noted that it could reduce the workload of teachers and provide immediate feedback to students. For example, Temara et al. \cite{temara2023maximizing} utilized a case study approach to examine and examine how ChatGPT can be used to gather useful reconnaissance data. ChatGPT is capable of generating a variety of intelligence related to specific targets, such as Internet Protocol (IP) address ranges, domain names, network topology, vendor technologies, ports, services, and even the operating systems used by the target. However, some users encountered issues with the accuracy of ChatGPT responses, as well as its limited ability to provide certain contextual information \cite{bishop2023computer}. 

\begin{figure}[h]
\centering

\tcbox[colframe=gray!30!black,
           colback=gray!20]{\includegraphics[width=.70\textwidth]{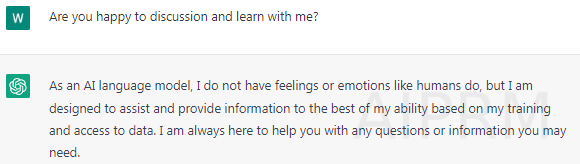}}
\caption{Example of emotion answer from ChatGPT.}
\label{fig:emotion}
\end{figure}

Additionally, there were some cases where ChatGPT provided alternative answers that contradicted previous responses given on the same topic.
In the context of exams, according to Teo Susnjak \cite{susnjak2022chatgpt}, The ability of ChatGPT to display analytical thinking abilities and produce very convincing text with little guidance makes it a possible danger to the credibility of online exams, especially in higher education contexts where such exams are becoming more common. Also, through interactions with ChatGPT, it became apparent that the chatbot is unable to express any emotions as shown in Figure \ref{fig:emotion}. As a result, there is a need to explore ways to make chatbots more human-like, not just in terms of their ability to provide thoughtful responses, but also in terms of their capacity to express emotions and exhibit a distinct personality. Other ChatGPT-based educational applications can be found in \cite{lund2023chatgpt,lee2023rise,wen2023future,lyerly2023utilizing,archibald2023chatgtp,zaabi2023review,elbanna2024exploring}.

\subsection{Customer service:} ChatGPT has been used to develop chatbots that can handle customer inquiries and support requests. These chatbots can understand natural language text and provide personalized responses, improving the overall customer experience and reducing the workload of customer service agents.

Also, in the business sector, chatGPTcan is able to enhance e-commerce via chat, finance and productivity. Some examples can be found in \cite{george2023review,george2023review}. Additionally, it can affect labor market \cite{zarifhonarvar2023economics}. It is expected that the use of AI may conduct workers to lose their jobs, especially those who perform repetitive tasks. This effect on the job market is linked to the fact that workers who are let go may not have the necessary skills to adapt to other types of work, leading to prolonged periods of unemployment. However, providing opportunities for training and acquiring new skills can help alleviate the negative impact of AI on the job market.

\subsection{Content creation:} ChatGPT has been used to generate high-quality content for websites, social media, and marketing campaigns \cite{li2023embracing}. The model can generate text in various formats, such as blog posts, product descriptions, and social media captions, and can adapt to different writing styles and tones. 

\begin{figure}[h]
\centering
{\includegraphics[width=0.80\textwidth]{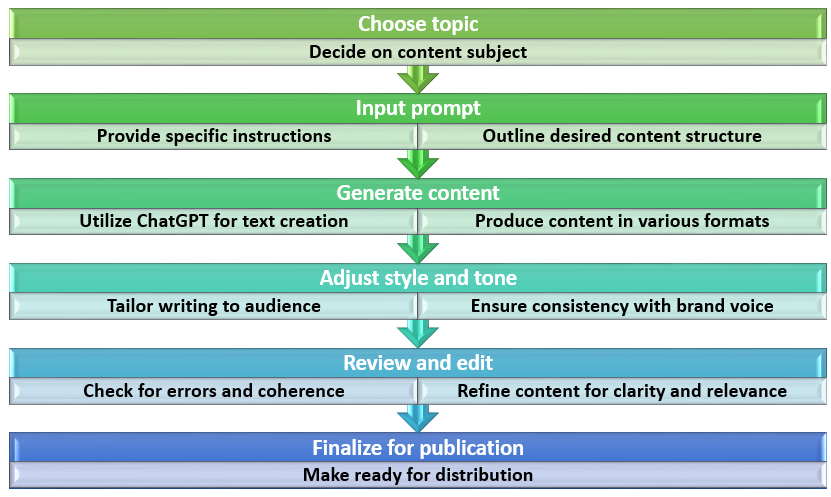}}
\caption{Steps in content generation with ChatGPT.}
\label{fig:content_creation}
\end{figure}

The input prompt serves as a comprehensive guide, offering specific instructions and outlining the desired structure for the content. Leveraging ChatGPT, content generation spans across various formats, ensuring versatility and adaptability to different platforms and audiences. Adjusting style and tone is imperative, tailoring the writing to resonate effectively with the intended readers while maintaining consistency with the brand voice. Upon content creation, meticulous review and editing processes are implemented, ensuring coherence, accuracy, and relevance. Refinements are made to enhance clarity, ultimately preparing the content for publication and distribution across relevant channels. The key steps of this process are given in Figure \ref{fig:content_creation}.

Regarding education content in the field of algebra, the authors in \cite{pardos2023learning} evaluated effectiveness of ChatGPT hints compared to those authored by human tutors in two algebra topic areas, Elementary Algebra and Intermediate Algebra, with 77 participants. They found that 70\% of the hints produced by ChatGPT passed manual quality checks and both human and ChatGPT hints resulted in positive learning gains. However, the gains were only statistically significant for hints created by human tutors. The study found that human-created hints resulted in significantly higher learning gains than ChatGPT hints in both topic areas, although ChatGPT participants in the Intermediate Algebra experiment were already at a high level and performed similarly to the control group at pre-test. Moreover, ChatGPT can also be used to generate social media content starting with textual content, and also multimedia content including images and videos using additional tools such as Canvas and Midjourney.

About plagiarism, Hamaweh \cite{halaweh2023chatgpt} states that If a student or academic writer employs ChatGPT to produce an essay but gives due credit to the ideas obtained through reverse searching, this would not constitute plagiarism. It is possible for students or researchers to plagiarize without using ChatGPT, but the difference now is that they can do it much more quickly. Nevertheless, this should not be viewed as an excuse to avoid utilizing ChatGPT. Ultimately, it is the responsibility of users, whether students or faculty, to use it responsibly by being properly informed and educated on how to do so. Without ChatGPT, one could still write a report or paper that incorporates ideas borrowed from others without being detected. Other ChatGPT-based content creation examples can be found in \cite{cao2023comprehensive,kirtania2023openai,htet2024chatgpt,zhang2024co,yu2024chatgpt,bringula2024chatgpt}.

\subsection{Language translation:} 
\begin{figure}[h]
\centering
{\includegraphics[width=0.80\textwidth]{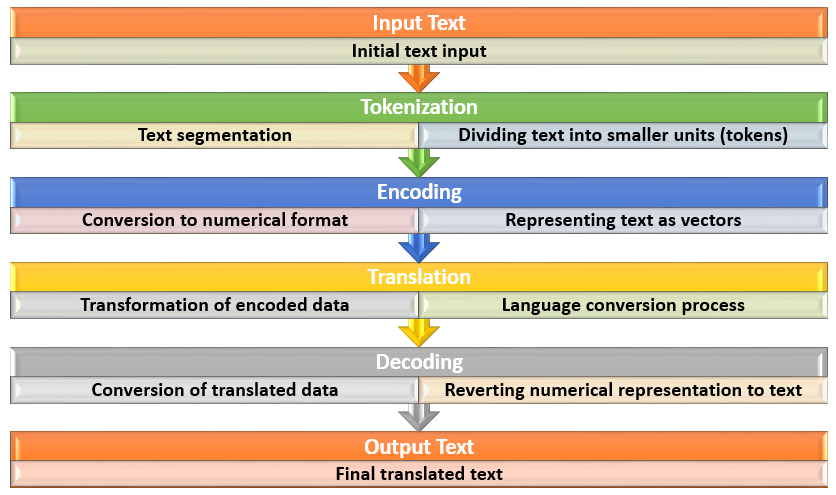}}
\caption{Language translation process overview.}
\label{fig:translationfig}
\end{figure}

ChatGPT has been used to develop language translation systems that can translate text between different languages with remarkable accuracy. These systems can understand the nuances of different languages and provide context-specific translations, improving communication between people from different cultures and backgrounds. Figure \ref{fig:translationfig} presents the language translation process overview using ChatGPT starting by transforming raw text into numerical representations which involves segmentation and tokenization, followed by encoding to create vectors. Translation occurs by processing encoded data, then decoding converts translated data back to text, resulting in the final translated output.

In \cite{jiao2023chatgpt}, the authors studied the performance of ChatGPT as a translator compared to Google, DeepL, and Tencent. Also, ChatGPT can be used for code translation between programming languages such as Java, Python, and others. This could be achieved by providing the code snippet in one language as input to the large language model, along with a prompt that specifies the desired output language. The model would then generate the equivalent code in the target language based on its understanding of the syntax and semantics of the input language and the prompt provided \cite{megahed2023generative}. However, it may not be as accurate or efficient as specialized code translation tools. It may also require specific prompts and inputs to accurately translate code.

According to Hamaweh \cite{halaweh2023chatgpt},  ChatGPT can prove to be highly advantageous by assisting in the quick generation of texts that would otherwise require significant time and effort by humans. Therefore, there is no need to worry about the efficiency of ChatGPT in generating, summarizing, translating, writing, and editing English texts.

Peng et al. \cite{peng2023towards} conducted research on ways to enhance ChatGPT translation proficiency by exploring three different viewpoints: temperature, task, and domain information. They also proposed two uncomplicated yet impactful prompts. Additionally, they conducted a comparison between ChatGPT and Google Translate's translation capabilities using various translation prompts.

\subsection{Entertainment:} 
\begin{figure}[h]
\centering
\tcbox[colframe=gray!30!black,
           colback=gray!20]{\includegraphics[width=0.60\textwidth]{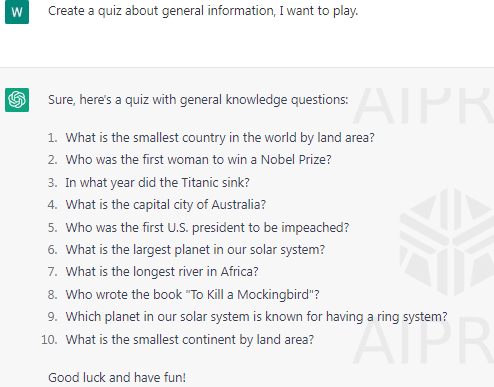}}
\caption{Example of a quiz generated by ChatGPT.}
\label{fig:quiz}
\end{figure}

ChatGPT has been used to develop chatbots that can simulate conversations with historical figures or fictional characters, providing a unique and engaging entertainment experience \cite{biswas2023role}. Moreover, ChatGPT can be used to generate interactive stories where the user can choose their own adventure by selecting different options presented to them by the model. This can provide a unique and engaging storytelling experience for the user. Also, ChatGPT can be used to create personality quizzes (See Figure \ref{fig:quiz}) where the model asks the user a series of questions to determine their personality traits or preferences. This can be a fun way for users to learn more about themselves and compare their results with others.

\subsection{Financial services:} The application of ChatGPT extends beyond casual conversation and into more specialized domains, such as finance. One notable application is the development of virtual financial advisors, leveraging the capabilities of ChatGPT to provide tailored investment recommendations and advice. \cite{chuma2023business,cao2023bridging}.

By harnessing ChatGPT's natural language processing abilities, these virtual financial advisors can engage in interactive dialogues with users, gathering information about their financial goals, risk preferences, and investment objectives \cite{oehler2024does}. This personalized approach allows the advisor to understand the unique needs of each individual and tailor its recommendations accordingly. Moreover, ChatGPT's deep learning capabilities enable it to analyze vast amounts of financial data, market trends, sentiment analysis, and investment strategies in real-time \cite{fatouros2023transforming,sudirjo2023application,rane2023role,sudirjo2023application}. By continuously learning from new information and market developments, the virtual financial advisor can provide up-to-date and informed recommendations to users, helping them navigate complex investment decisions with confidence.

\begin{figure}[h]
\centering
\includegraphics[width=0.99\textwidth]{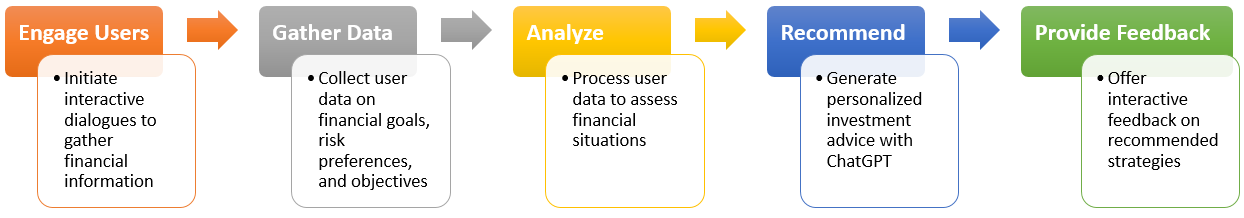}
\caption{Steps of virtual financial advisor implementation using ChatGPT.}
\label{fig:virtual_advisor}
\end{figure}

Implementing virtual financial advisors with ChatGPT entails engaging users in interactive dialogues to gather comprehensive financial information, including their goals, preferences, and investment objectives. Once gathered, the data undergoes meticulous analysis to assess individual financial situations accurately \cite{huang2023chatgpt,li2024can}. Leveraging ChatGPT's natural language processing capabilities, personalized investment recommendations are generated, taking into account factors such as risk tolerance and market conditions. Finally, users receive detailed and interactive feedback on the recommended strategies, ensuring transparency and understanding. This process optimizes the use of ChatGPT to deliver tailored financial guidance effectively, fostering trust and satisfaction among users. 
Buadottir et al. \cite{buadottir2023kira} presents a financial advisory called Kira via text messages. It utilizes ChatGPT to provide responses, accesses users' financial data through the Plaid platform, and communicates through SMS messages using Twilio. To address privacy concerns, we developed an algorithm to encrypt user data before sending it to ChatGPT and decrypt it for responses. This showcases the integration of advanced natural language AI with other tools to create a comprehensive and user-friendly application. Figure \ref{fig:virtual_advisor} summarizes the key steps of virtual financial advisor implementation using ChatGPT.

Additionally, the integration of business AI decision-making further enhances the capabilities of these virtual financial advisors \cite{jusman2023application}. By incorporating advanced algorithms and predictive analytics, they can assess the potential risks and rewards associated with different investment options, optimize portfolio allocations, and even automate trading decisions based on predefined criteria.

These are just a few examples of the many applications of ChatGPT in various industries and fields \cite{dowling2023chatgpt,zaremba2023chatgpt}. As technology continues to evolve and improve, it is expected to play an increasingly important role in our daily lives, transforming the way we interact with machines and each other.

\subsection{Atmospheric science:}
ChatGPT has the capability to significantly contribute to enhancing our comprehension of climate change and refining the precision of climate forecasts. Various uses of ChatGPT can assist climate research, such as in interpreting and analyzing data, generating scenarios, evaluating models, and parameterizing models \cite{biswas2023potential}. 
By engaging with researchers, students, and enthusiasts alike, ChatGPT facilitates inquiries into meteorology, climate dynamics, and atmospheric physics. Through its sophisticated language processing capabilities, it can comprehend complex queries, ranging from inquiries about atmospheric phenomena to discussions on climate modeling techniques \cite{zhu2023chatgpt,agathokleous2023use}. Furthermore, ChatGPT's ability to recognize intents and manage conversations enables it to guide users through intricate concepts and datasets, fostering deeper understanding and exploration within the field. ChatGPT additionally possesses the capability to identify and map flood-prone areas in real-time, showcasing its potential to revolutionize fields such as environmental monitoring and disaster management by aiding in the detection of floods \cite{kumbam2024floodlense}.

By utilizing a variety of data sources, this technology provides researchers and policymakers with a powerful tool to generate and evaluate a wide range of climate scenarios. This process significantly improves the precision and reliability of climate projections. By incorporating diverse data inputs, including meteorological data, satellite observations, historical climate data, and various climate models, this technology enables a comprehensive understanding of the complexities of climate dynamics. Moreover, it assists in pinpointing areas susceptible to risks and crafting specific plans for enhancing the resilience of infrastructure, preparing for disasters, and fostering community involvement. \cite{rane2024contribution}. This holistic approach allows for a more thorough examination of potential future climate trends and their potential impacts. As a result, decision-makers can make more informed choices regarding climate adaptation and mitigation strategies. Overall, the integration of diverse data inputs along with ChatGPT tools enhances the accuracy and robustness of climate projections, aiding in the development of effective climate policies and initiatives.

\subsection{Chatbots:}
Chatbots are one of the most common applications of ChatGPT. A chatbot is an AI-powered computer program designed to simulate human conversation, usually through text or voice interactions \cite{tlili2023if,alzu2024exploring}. A comprehensive comparison between traditional chatbots and ChatGPT chatbot can be found in \cite{panda2023exploring}.

Chatbots can be integrated into various platforms, such as websites, social media, messaging apps, and voice assistants, to provide personalized customer support, automate repetitive tasks, and engage with users. Recently, many browser extensions have been introduced to facilitate the use of ChatGPT in the web, emails, and also through vocal requests such as ChatGPT Writer, ChatGPT for google, and Youtube summary with ChatGPT. 25 best ChatGPT Chrome extensions can be found in \cite{WinNT}. These extensions are already provided in Bing browser of Microsoft. 
The key steps for conversational AI development to build a a Chatbot with ChatGPT are: 

\textbf{Data Collection:} Gather a diverse range of conversational data to train the ChatGPT model, encompassing various topics and conversation styles.

\textbf{Preprocessing:} Clean and preprocess the collected data, including tokenization, removing noise, and formatting for compatibility with the model.

\textbf{Model Training:} Train the ChatGPT model on the preprocessed data, utilizing techniques such as fine-tuning on conversational datasets to enhance its ability to generate coherent responses.

\textbf{Integration:} Integrate the trained model into the chatbot framework to process user inputs and generate responses in real-time. This integration ensures seamless interaction between users and the chatbot, enabling it to understand queries and provide appropriate answers instantly. By embedding the model within the framework, the chatbot can analyze contextual information and deliver relevant responses promptly. This real-time capability enhances user satisfaction and engagement with the chatbot experience.

\textbf{Deployment:} 
Deploy the chatbot across various platforms, including websites, messaging apps, and voice interfaces, to ensure its accessibility to users. This involves adapting the chatbot's interface and functionality to each platform's specifications and user interface guidelines. By seamlessly integrating the chatbot into these diverse platforms, users can interact with it effortlessly, regardless of their preferred communication channel.

\textbf{Evaluation and Improvement:} Continuously evaluate the chatbot's performance through user feedback and monitoring metrics like response coherence and relevance. Iterate on the model and update it regularly to enhance its conversational abilities.

\begin{figure}[h]
\centering
\includegraphics[width=0.66\textwidth]{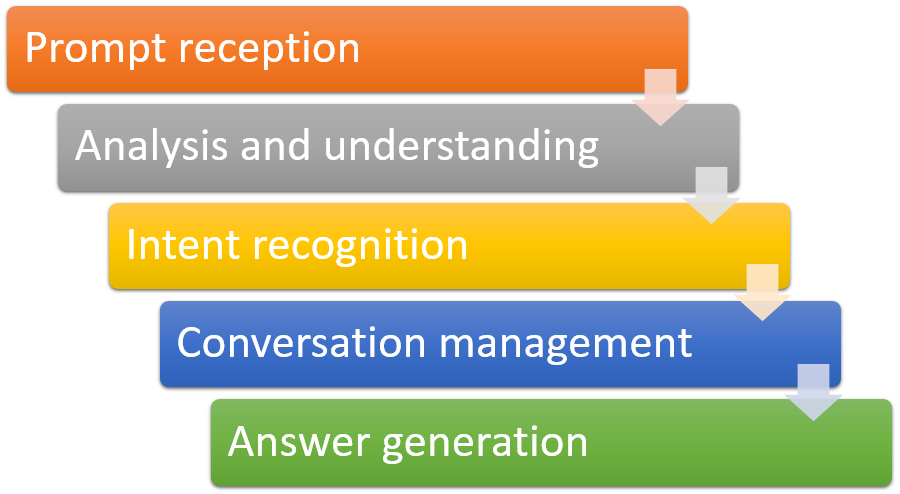}
\caption{ChatGPT chatbot steps.}
\label{fig:chatbotsteps}
\end{figure}

Figure \ref{fig:chatbotsteps} presents the ChatGPT chatbot steps after its deployment. Upon receiving the prompt, ChatGPT initiates by analyzing and understanding the input. It then proceeds to recognize the user's intent, enabling effective conversation management. Utilizing this understanding, it generates coherent and relevant answers, ensuring a seamless and engaging interaction with the user. This cyclical process forms the backbone of ChatGPT's conversational abilities, allowing it to adapt and respond dynamically to various queries and topics.

One of the main advantages of using ChatGPT in chatbots is that it allows for more natural and human-like conversations. With its advanced language processing capabilities and vast knowledge base, ChatGPT can understand complex queries, intent and respond with appropriate and relevant information \cite{alshurafat2023usefulness,he2023can}. This makes chatbots more efficient and effective in handling customer inquiries and support requests, while also improving the overall customer experience.

Chatbots powered by ChatGPT can be used in a variety of industries and sectors. For example, in healthcare, chatbots can be used to provide medical advice, medication reminders, and mental health support. Other examples can be found in \cite{chow2023impact,shahsavar2023role,loh2023chatgpt}. In finance, chatbots can be used to help customers manage their accounts, make payments, and get investment advice. In e-commerce, chatbots can be used to provide personalized product recommendations, track orders, and handle returns \cite{dowling2023chatgpt}.
One of the challenges in using ChatGPT in chatbots is ensuring that the responses generated by the model are accurate and relevant \cite{king2023conversation}. This requires a large and diverse training dataset, as well as careful monitoring and quality assurance measures to ensure that the chatbot is providing helpful and accurate information.
Despite these challenges, chatbots powered by ChatGPT are becoming increasingly popular in various industries, and are expected to play a significant role in transforming the way businesses interact with their customers in the future.

During the first month following its launch, the author in \cite{taecharungroj2023can} gathered tweets discussing ChatGPT, a novel AI chatbot. By analyzing 233,914 tweets in English with the latent dirichlet allocation topic modeling algorithm, the author aimed to uncover the abilities of ChatGPT. The findings indicated that the discussions surrounding ChatGPT on Twitter primarily centered around news, technology, and reactions. Also, Salah et al. \cite{salah2023chatting} conducted a research project that explored the connections among several factors, including trust in ChatGPT, how users perceive ChatGPT, stereotypes associated with ChatGPT, as well as two psychological outcomes: self-esteem and psychological well-being.

\subsection{Customization and personalization:}
In today's digital landscape, organizations strive to engage users effectively, whether it's through websites, mobile applications, or other digital platforms \cite{lee2024establishing}. One potent tool in this endeavor is the implementation of FAQ chatbots, which serve as virtual assistants, guiding users through common queries and providing instant support. Among the myriad options available, ChatGPT stands out for its remarkable customization and personalization capabilities, offering organizations a powerful means to elevate their user experience. In the following, we present some examples of customization and personalization services using ChatGPT chatbots.

\textbf{Customization: tailoring chatbots to reflect brand identiy:}

One of the key advantages of ChatGPT lies in its ability to be tailored precisely to an organization's needs and branding. This level of customization enables organizations to create a chatbot experience that seamlessly aligns with their identity, ensuring a consistent brand voice and aesthetic throughout user interactions. For instance, companies can customize ChatGPT to match their logo, colors, and tone of communication, reinforcing brand recognition and loyalty among users. Moreover, organizations can leverage ChatGPT's customization features to provide accurate responses to user inquiries \cite{agarwal2022descriptive}. By fine-tuning the chatbot's knowledge base and algorithms, businesses can ensure that users receive prompt and helpful answers to their questions, thereby enhancing overall satisfaction and trust in the brand.

Furthermore, ChatGPT's customization capabilities extend beyond just aesthetics and responses. Organizations can integrate the chatbot with existing systems and databases, allowing for seamless access to relevant information and resources. This integration streamlines workflows and enhances efficiency, enabling the chatbot to serve as a valuable asset in various business processes.

\textbf{Personalization: tailoring responses to individual preferences:}
In addition to customization, ChatGPT excels in personalizing responses based on user preferences and past interactions such as in business and tourism services \cite{harahap2023use,george2023review,remountakis2023using}. This level of personalization holds immense potential for enhancing the user experience, making interactions with the chatbot more intuitive and engaging. 
For example, ChatGPT can analyze user behavior and interactions to tailor responses accordingly. If a user frequently asks about a particular product or service, the chatbot can proactively provide information and recommendations related to that topic, anticipating the user's needs and preferences.

Moreover, ChatGPT can adjust the conversation flow based on previous interactions, ensuring a seamless and contextually relevant dialogue especially in e-commerce \cite{el2023sentiment}. By considering the user's journey and the context of their queries, the chatbot can guide users through complex processes or provide assistance in a manner that feels natural and intuitive. 
Furthermore, ChatGPT can leverage user-specific information, such as location or purchase history, to deliver more personalized responses. For instance, if a user inquires about nearby stores or events, the chatbot can provide tailored recommendations based on their geographical location, enhancing the relevance and usefulness of the interaction.

Table \ref{tab:pers_cust} presents a detailed table outlines various aspects of customization and personalization in FAQ chatbots, providing interpretation, examples, advantages, and challenges associated with each aspect. This table summarizes key aspects of customization and personalization in FAQ chatbots: brand alignment, accuracy, integration, personalized responses, proactive assistance, and emotion recognition. It highlights how each aspect contributes to user experience. Challenges include ensuring accuracy, compatibility, and data security. Overall, these features enhance user engagement, trust, and satisfaction by tailoring responses, anticipating needs, and recognizing user emotions.

\begin{sidewaystable}[htbp]
\centering
\caption{Customization and personalization aspects in FAQ Chatbots}
\label{tab:pers_cust}
\begin{tabular}{|m{2.5cm}|m{4.5cm}|m{4.5cm}|m{4.5cm}|m{4.5cm}|}
\hline
\textbf{Aspect}            & \textbf{Interpretation}                                                                                                           & \textbf{Examples}                                                                                                                                                                                       & \textbf{Advantages}                                                                                                                & \textbf{Challenges}                                                                                                                                                  \\ \hline
Brand Alignment             & Customizing the chatbot to align with the organization's branding, including colors, logos, and tone of voice.               & - Using company colors and logo in the chatbot interface. - Tailoring the chatbot's language and tone to match the brand's identity.                                                                  & - Reinforces brand identity and consistency. - Enhances brand recognition among users.                                             & - Ensuring the chatbot's branding doesn't overshadow its functionality. - Maintaining consistency across different platforms and channels.                                 \\ \hline
Accuracy                    & Ensuring the chatbot provides accurate and relevant responses to user queries.                                                  & - Training the chatbot on a comprehensive dataset of frequently asked questions. - Regularly updating the chatbot's knowledge base to reflect changes or new information.                                   & - Builds user trust and confidence in the chatbot's capabilities. - Reduces frustration by providing correct and helpful information. & - Ensuring the chatbot's responses remain accurate as information evolves. - Handling ambiguous or complex queries accurately.                                           \\ \hline
Integration                 & Seamlessly integrating the chatbot with existing systems or platforms used by the organization.                                & - Integrating the chatbot with customer relationship management (CRM) software to access user data. - Connecting the chatbot with e-commerce platforms for product recommendations and purchases. & - Streamlines workflows by automating tasks and processes. - Provides a unified user experience across different platforms.          & - Compatibility issues with legacy systems or proprietary software. - Ensuring data privacy and security when integrating with external systems.                       \\ \hline
Personalized Responses      & Tailoring responses based on user preferences, previous interactions, and contextual information.                               & - Recommending products or services based on past purchases or browsing history. - Addressing users by name and acknowledging previous interactions.                                                    & - Increases user engagement and satisfaction by providing relevant and timely information. - Enhances the perception of the chatbot as attentive and user-focused.     & - Balancing personalization with privacy concerns and data protection regulations. - Ensuring accuracy and relevance in personalized recommendations and responses.           \\ \hline
Proactive Assistance        & Anticipating user needs and providing assistance or information before users ask for it.                                      & - Offering assistance when users linger on a particular webpage or feature. - Sending proactive notifications or reminders based on user preferences.                                                  & - Improves user experience by reducing the need for users to initiate interactions. - Increases efficiency by addressing potential issues before they escalate.       & - Avoiding over-reliance on proactive assistance, ensuring users still feel in control. - Striking the right balance between helpfulness and intrusiveness.                 \\ \hline
Emotion Recognition        & Detecting and responding to user emotions through language cues or sentiment analysis.                                        & - Adjusting responses based on the tone or sentiment of user input. - Offering empathy and support in response to expressions of frustration or dissatisfaction.                                           & - Enhances the chatbot's ability to empathize and connect with users on a deeper level. - Improves user satisfaction by acknowledging and addressing emotional needs.  & - Ensuring accuracy in emotion detection, avoiding misinterpretation of user intent. - Handling sensitive emotional responses appropriately and sensitively.             \\ \hline
\end{tabular}
\end{sidewaystable}

\subsection{ChatGPT for computer science developers and coding:}
ChatGPT can also be useful for computer science developers and coding enthusiasts. One of the main applications of ChatGPT in this field is code generation. With its natural language processing capabilities, ChatGPT can understand human-readable descriptions of programming tasks and generate code to solve them. Utilizing ChatGPT can also be advantageous in creating queries for extensive systematic reviews that have a high level of accuracy as shown in \cite{wang2023can}.

\begin{figure}[h]
\centering
\includegraphics[width=0.65\textwidth]{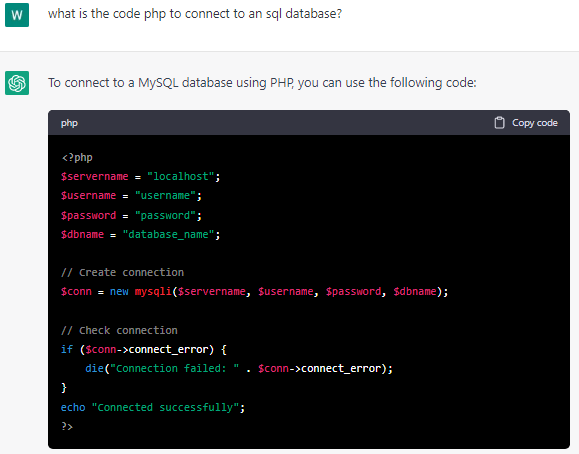}
\caption{Example of PHP code written by ChatGPT.}
\label{fig:php}
\end{figure}

For example, OpenAI has released a language model called Codex, which is based on GPT technology and trained on a massive dataset of code. Codex can understand natural language descriptions of coding tasks and generate code to solve them in various programming languages, such as Python, Java, and C++, and PHP as shown in Figure \ref{fig:php}. This can save developers a significant amount of time and effort in writing code from scratch, especially for repetitive or routine tasks.

Another application of ChatGPT in computer science is natural language programming. This involves creating programming languages that can be written and executed using natural language text instead of traditional programming syntax. With ChatGPT advanced language processing capabilities, it can be used to develop more intuitive and accessible natural language programming interfaces, making coding more accessible to people with different backgrounds and skill levels. ChatGPT can also be used to generate documentation and tutorials for programming languages and tools \cite{avila2023chatgpt}. With its ability to generate high-quality text, ChatGPT can help developers create clear and concise documentation that is easy to understand and follow. Generally, ChatGPT has the potential to revolutionize the way developers write and interact with code, making programming more accessible, efficient, and intuitive.
To sum up, ChatGPT can enhance research and scholarship in academia through several ways, it can:

\begin{itemize}
    \item Assist researchers in identifying relevant literature by generating summaries of articles or providing a list of relevant papers based on a given topic or keyword.
    \item Generate text in a specific style or tone, making it easy for researchers to produce draft versions of research papers, grant proposals, and other written materials.
    \item Help researchers analyze large amounts of text data, such as social media posts or news articles, by providing insights and identifying patterns in the data.
    \item Be used for machine translation, making research materials accessible in multiple languages.
    \item Summarize scientific papers, reports, or other documents, making it easier for researchers to stay up-to-date with the latest developments in their field.
    \item Answer domain-specific questions, making it a powerful tool for scholars to find answers quickly and efficiently.
\end{itemize}
All these capabilities of ChatGPT can assist researchers in saving time and effort, enabling them to concentrate on the more analytical and creative aspects of their work.
However, when using ChatGPT for academic writing and scientific research, it's essential to consider certain limitations that might undermine the research's quality. For instance, the utilization of ChatGPT has often been criticized for producing superficial, inaccurate, or erroneous content in scientific writing \cite{stokel2023chatgpt,marchandot2023chatgpt,zhou2024chatgpt,khalifa2024artificial,margetts2024use}. 

The ethical concerns associated with ChatGPT usage involve the potential for bias resulting from the training datasets and the possibility of plagiarism. Additionally, the lack of transparency in the process of content generation has led to ChatGPT being labeled as a "black box" technology \cite{lubowitz2023chatgpt,biswas2023chatgpt,lund2023chatting,liebrenz2023generating}.

\section{Advantages of ChatGPT in natural language processing}
\label{sec:ad}
The advantages of ChatGPT in natural language processing are numerous and significant. Here are some examples:

\begin{itemize}[label=\textbullet]
\item \textbf{Improved language understanding:} ChatGPT is capable of understanding natural language text better than previous language models. It can understand complex sentence structures, idiomatic expressions, and even sarcasm, making it more effective at processing natural language.

\item \textbf{Better response quality:} With its large-scale training and ability to generate human-like responses, ChatGPT can produce high-quality responses to natural language queries. This is especially important in chatbots and customer service applications where accuracy and relevance are crucial.

\item \textbf{Increased efficiency:} ChatGPT can help streamline natural language processing tasks, such as language translation, text summarization, and sentiment analysis. By automating these tasks, ChatGPT can save time and reduce the need for manual labor.

\item \textbf{Language adaptability:} ChatGPT can be trained on data from different languages and dialects, making it adaptable to various linguistic contexts. This means that it can provide accurate and relevant responses in multiple languages, which is essential in a globalized world.

\item \textbf{Personalization:} ChatGPT can learn and adapt to individual users' language preferences, writing style, and context. This means that it can generate personalized responses that are tailored to the user's needs, leading to a better user experience.

\item \textbf{Scalability:} ChatGPT can process vast amounts of natural language data quickly and efficiently. This makes it suitable for handling large-scale language processing tasks, such as social media monitoring or content analysis.

\item \textbf{Accessibility:} ChatGPT can be used by people with varying levels of language proficiency, including non-native speakers and people with disabilities. It can help improve communication and accessibility for people who struggle with traditional written or spoken language. ChatGPT can also be integrated with other technologies such as voice recognition and image recognition, enabling it to provide more comprehensive services.

\item \textbf{Fine-tuning:} ChatGPT can be fine-tuned for diverse tasks, showcasing its adaptability in handling various domains such as customer support, content generation, and automation of text-based functions. The process of fine-tuning, including data preparation, model selection, hyperparameter tuning, and deployment strategies, highlights ChatGPT's versatility and potential for transformative applications in specialized domains. Examples of fine-tuned ChatGPT LLM in automatic scoring and recommendation can be found in \cite{lv2023full,latif2024fine,li2023exploring}
\end{itemize}
Some real-world examples of ChatGPT advantages in natural language processing include:

\begin{itemize}[label=\textbullet]
\item \textbf{Language Translation:} ChatGPT can be used to translate text from one language to another, providing accurate and relevant translations that are similar to human translations. This can help businesses and organizations communicate effectively with customers and partners in different countries and regions.

\item \textbf{Text Summarization:} ChatGPT can be used to summarize lengthy text passages, such as news articles or research papers, into shorter and more digestible summaries. This can help users save time and improve their reading comprehension.

\item \textbf{Sentiment Analysis:} ChatGPT can be used to analyze text data, such as social media posts or customer reviews, to determine the sentiment or emotion behind the text \cite{hariri2024sentiment}. This can help businesses and organizations monitor their reputation and customer feedback.

\item \textbf{Prediction:} ChatGPT can be used in prediction scenarios based on input data. For example, in scientific writing, it can predict citation counts using scientific abstracts as shown in \cite{de2024can}. This study was based on sentiment analysis, and linguistic complexity analysis \cite{hassan2020predicting}. Similarly, Hu et al. \cite{hu2021analysis} results showed that abstracts from papers with high citation counts exhibited a more intricate vocabulary, longer sentences, and more complex syntactic structures, resulting in lower readability compared to abstracts from uncited papers.
\end{itemize}
The aforementioned advantages of ChatGPT in natural language processing make it a powerful tool for improving communication, enhancing user experience, and automating language processing tasks in various industries and applications.

\section{Limitations and potential challenges}
\label{sec:dis}
While ChatGPT has numerous advantages in natural language processing and has been applied successfully in various real-world scenarios, there are still limitations and potential challenges to consider. These limitations and challenges are important to understand to ensure the effective and ethical use of ChatGPT in various applications. In this section, we will explore some of the potential limitations and challenges of ChatGPT.
\subsection{Bias and harmful language patterns}
One of the potential limitations and challenges of ChatGPT is the issue of bias and harmful language patterns. ChatGPT is trained on vast amounts of text data, including online forums, social media posts, and news articles, which can contain biased language and harmful stereotypes. As a result, ChatGPT can reproduce these biases and harmful language patterns in its responses. For example, if ChatGPT is trained on a dataset that contains sexist or racist language, it may produce responses that reflect these biases. This can be problematic in applications such as chatbots, where these responses can reinforce harmful stereotypes and perpetuate discrimination \cite{mcgee2023chat}.

To address this issue, researchers and developers have proposed various approaches, including pre-processing and filtering the training data to remove biased or harmful language, incorporating diverse perspectives and sources in the training data, and implementing bias detection and mitigation techniques in the ChatGPT model.

Another related issue is the potential for ChatGPT to generate hate speech or other harmful language patterns. In some cases, users may intentionally input hate speech or other harmful language into the system, and ChatGPT may reproduce these harmful patterns in its responses. To address this issue, developers can implement content moderation techniques and filters to detect and prevent hate speech and other harmful language patterns from being generated. To sum up, addressing the issue of bias and harmful language patterns in ChatGPT is crucial to ensure that its applications are ethical and socially responsible. Ongoing research and development in this area are essential to mitigate these challenges and maximize the potential benefits of ChatGPT in natural language processing.
\subsection{Limited ability to understand context}
Another potential limitation of ChatGPT is its limited ability to understand context. ChatGPT is trained on large amounts of text data, but it does not have the same level of understanding of context as humans do \cite{cascella2023evaluating}. This means that ChatGPT may struggle to understand the meaning of a message or conversation in the same way that a human would.
For example, if a user asks a question that is ambiguous or unclear, ChatGPT may not be able to fully understand the context and provide an accurate response. Similarly, if a user uses sarcasm or irony in their message, ChatGPT may not be able to detect the tone and provide an appropriate response.
To address this limitation, developers can implement techniques such as context modeling and entity recognition to enhance understanding of context. This involves analyzing the content and context of the message or conversation to identify important entities, such as people, places, or events, and use this information to generate a more accurate response.

Despite continuous efforts to enhance ChatGPT's comprehension of context, there remains a significant amount of work to be done in this domain. This limitation can potentially affect the accuracy and utility of ChatGPT in specific applications. While ChatGPT excels in many tasks, its understanding of context may still fall short in nuanced or complex scenarios. Developers need to be mindful of these limitations when designing and implementing applications that rely on ChatGPT's capabilities. Understanding the boundaries of ChatGPT's contextual comprehension is crucial for mitigating potential errors or misunderstandings in its responses \cite{huang2023chatgpt}. 

By acknowledging these constraints and designing systems that work within them, developers can maximize the effectiveness and reliability of ChatGPT in various contexts, including environmental monitoring and disaster management applications like flood detection. Continuous research and development efforts aimed at improving ChatGPT's contextual understanding will be vital in further enhancing its performance and expanding its range of applications in the future.
\subsection{Need for large amounts of training data}
ChatGPT requires large amounts of training data to achieve its state-of-the-art performance in natural language processing. This means that developers need to have access to vast quantities of high-quality text data in order to train the model effectively.
For example, the original GPT model from OpenAI was trained on a dataset of over 40 GB of text data, while the largest version of GPT-3.5 was trained on a dataset of over 570 GB of text data. This amount of data can be difficult to obtain, particularly for smaller organizations or those without access to large amounts of text data.
In addition to the quantity of data, the quality of the data is also important. The training data must be representative of the language and domains that the ChatGPT model will be used for, and it must be free from biases and errors that could impact the accuracy of the model.

\begin{figure}[h]
\centering
\includegraphics[width=0.99\textwidth]{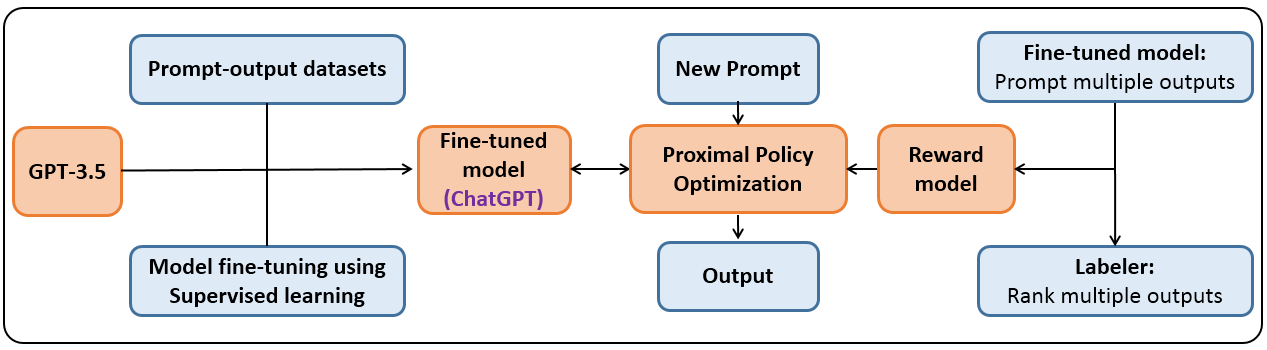}
\caption{GPT-3.5 boosted to ChatGPT using human prompts, Reinforcement Learning, and fine-tuning techniques \cite{zhou2023comprehensive}.}
\label{fig:finetune}
\end{figure}

The need for large amounts of training data can also pose challenges for fine-tuning ChatGPT for specific applications \cite{zhou2023comprehensive}. Fine-tuning involves further training the pre-trained ChatGPT model on a smaller dataset of domain-specific text data to improve its performance on specific tasks. However, collecting and labeling this additional training data can be time-consuming and expensive. Figure \ref{fig:finetune} presents how GPT-3.5 can be boosted to ChatGPT using human prompts and Reinforcement Learning.
Therefore, the need for large amounts of high-quality training data is a potential limitation of ChatGPT that must be considered in its development and application. Developers must carefully consider the availability and quality of training data when designing and implementing ChatGPT for specific applications. Other than Reinforcement learning, there are other types including supervised learning, semi-supervised learning, weakly-supervised learning \cite{mallick2023analyzing}, and self-supervised learning \cite{zhang2023complete}. 

Generative adversarial networks are mainly used with self-supervised learning. This method involves a collaboration between two deep models (generator and discrimination) with the goal of enhancing each other's capabilities. For instance, one AI generates data that closely resemble real ones, while the other model distinguishes between authentic and fake data. However, in the case of the "Reflexion" technique, GPT-4 performs both the role of the writer and editor, improving its own output.


\subsection{Cybersecurity}
Large language models like ChatGPT offer a myriad of positive benefits, one of which is their capacity to educate individuals who may not have extensive cybersecurity expertise \cite{al2023chatgpt,alawida2023comprehensive}. These models serve as accessible and intuitive tools for explaining complex cybersecurity concepts in plain language, thereby democratizing knowledge and fostering a deeper understanding of cybersecurity issues among a broader audience.

By leveraging ChatGPT, non-cybersecurity experts can gain insights into various aspects of cybersecurity, ranging from fundamental concepts like encryption and authentication to more advanced topics such as threat detection and mitigation strategies. Through interactive conversations with ChatGPT, users can ask questions, seek explanations, and receive informative responses tailored to their level of understanding.

Moreover, ChatGPT can simulate real-world cybersecurity scenarios, allowing users to explore the implications of different security measures and vulnerabilities in a safe and controlled environment. This hands-on learning experience enables non-experts to grasp the significance of cybersecurity practices and encourages proactive engagement in safeguarding digital assets and privacy.

Additionally, ChatGPT can act as a virtual cybersecurity tutor, guiding users through educational resources, tutorials, and best practices. By providing personalized recommendations and actionable insights, these models empower individuals to enhance their cyber hygiene practices and contribute to a more secure online ecosystem.

The deployment of ChatGPT chatbots, however, presents considerable cybersecurity risks that require attention and resolution. Moreover, a user's device may become infected with harmful software if they receive a malicious link or file through chatGPT \cite{sebastian2023chatgpt,o2023chatgpt}. In the past, cybercriminals were often constrained in their ability to carry out complex attacks because they required skills in coding and scripting, which involved writing new malware code and password-cracking scripts. However, there are concerns that ChatGPT could reduce the barriers to becoming a script kiddie or a cybercriminal, as it could potentially be used to generate computer malware code or password-cracking software. This could allow individuals with limited coding skills to engage in malicious activities that were previously only accessible to more skilled hackers. In \cite{al2023chatgpt}, The authors showcase in a case study how a ChatGPT can be utilized to devise and implement False data injection attacks on critical infrastructure, including industrial control systems.

Security analytics play a crucial role in detecting potential cyber threats by analyzing vast quantities of data to identify anomalies and patterns that may indicate impending attacks. These data sources include security tools, personal devices, network hardware, and server logs, all of which produce substantial amounts of information essential for threat detection and prevention. the study in \cite{sharma2023impact} explores the utilization of Big Data analytics and artificial intelligence platforms such as ChatGPT in addressing cybersecurity issues. It delves into the capabilities of current AI and data analytic technologies and their potential to bolster cybersecurity measures.

On the other hand, protecting medical information is crucial in healthcare because it contains sensitive and personal data, such as patient information, medical history, and health records. According to Mijwil et al. \cite{mijwil2023chatgpt}, the ways to  safeguard medical information consist of:

\begin{itemize}
    \item \textbf{Encryption and access controls:} These methods involve encoding medical data and implementing strict access controls to prevent unauthorized access to medical information.
    \item \textbf{Regular software updates:} Keeping software and systems up-to-date with the latest security patches can help protect against known vulnerabilities.
    \item \textbf{Network security:} Implementing firewalls, intrusion detection and prevention systems, and other network security measures can help protect against cyberattacks.
    \item \textbf{Risk management:} Regularly assessing and managing potential security risks can help healthcare organizations identify and address vulnerabilities before they can be exploited.
    \item \textbf{Compliance and employee education:} Adhering to industry regulations, and regularly educating and training employees on security best practices can help ensure that medical information is handled and protected in accordance with legal and ethical standards.
\end{itemize}

To ensure ethical handling of medical data, encryption, access controls, regular software updates, network security, risk management, compliance with regulations, and employee education are vital steps. These measures collectively protect patient information, uphold legal standards, and promote ethical conduct in healthcare practices.

\subsection{Response quality}
\begin{figure}[h]
\centering
\tcbox[colframe=gray!30!black,
           colback=gray!20]{\includegraphics[width=0.70\textwidth]{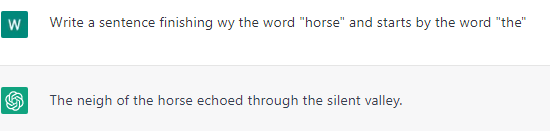}}
\caption{Wrong answer generated by ChatGPT.}
\label{fig:incorrect}
\end{figure}

It was stated that the virtual assistant ChatGPT can sometimes make mistakes and has limited information available (as of 2021 according to OpenAI). While most of the time ChatGPT provides reasonable and reliable responses, there are times when it may give incorrect or misleading information as shown in Figure \ref{fig:incorrect}, where it misses the sequence of the words, and can't provide a correct sentence according to the user request. This means that the quality of ChatGPT output is acceptable but could still be improved. One participant, a programmer, gave an example of ChatGPT generating incorrect code that did not work properly in programming software. However, some participants still praised ChatGPT as an efficient virtual assistant for creating knowledge and products due to its relatively low number of errors. some examples can be found in \cite{tlili2023if}.

Also, one of the major concerns of ChatPGT is the potential for false information to be disseminated through ChatGPT. There is a risk that ChatGPT could be used to create and spread false information or propaganda because it has the ability to produce text that resembles human writing, which may make it appear more reliable and trustworthy than content created by automated systems or bots \cite{sebastian2023chatgpt}.

\section{Ethical considerations when using ChatGPT}
\label{sec:ethi}
As with any advanced technology, there are ethical considerations that must be taken into account when using ChatGPT. These considerations include issues related to privacy, bias, science, and harmful language patterns \cite{robinson2023cost}.
One major ethical concern is the potential for ChatGPT to be used for malicious purposes, such as generating fake news or deepfakes. ChatGPT can be trained to generate text that is difficult to distinguish from human-generated text, which could be used to spread misinformation or manipulate public opinion. This highlights the importance of responsible use and regulation of ChatGPT and similar technologies.

Another ethical consideration is the potential for bias in ChatGPT output \cite{zhuo2023exploring}. ChatGPT is trained on large amounts of text data, which can reflect biases present in the data \cite{cong2024systematic}. This can lead to biased or unfair outputs, particularly in sensitive areas such as healthcare or criminal justice. Developers must carefully consider the potential for bias and take steps to mitigate it, such as using diverse and representative training data and evaluating the model's output for fairness.
Harmful language patterns are another ethical concern related to ChatGPT. ChatGPT may inadvertently generate text that contains harmful or offensive language, such as hate speech or profanity. Developers must consider the potential impact of ChatGPT output on users and take steps to prevent the generation of harmful or offensive language patterns.

Privacy is also an ethical consideration when using ChatGPT. ChatGPT may collect and store user data, such as chat logs, which could be used for malicious purposes or shared without user consent \cite{cardoso2023we}. Developers must ensure that user data is protected and handled responsibly in accordance with relevant privacy regulations.
Overall, there are a range of ethical considerations that must be taken into account when using ChatGPT. Developers and users of the technology must be aware of these considerations and take steps to mitigate any potential negative impacts. Responsible use and development of ChatGPT can help to ensure that the technology is used in a way that benefits society as a whole. We summarize the ethics issues in the following three points

\subsection{Fairness and bias in training data}
Fairness and bias in training data are critical considerations when using ChatGPT due to its reliance on large text datasets. These datasets may inadvertently capture and propagate biases present in society, leading to biased or unfair outputs, especially in sensitive domains like healthcare or criminal justice.

To address these challenges, developers must employ several strategies:
\begin{itemize}
    \item \textbf{Diverse and representative training data:} Ensuring that the training data encompasses diverse perspectives and demographics can help mitigate biases \cite{kuhlman2020no}. This includes considering factors such as age, gender, race, ethnicity, and socioeconomic background in the dataset.

    \item \textbf{Evaluation for fairness:} Developers should evaluate the model's outputs to identify potential biases systematically. This evaluation involves analyzing the model's responses across different demographic groups to detect any disparities or discriminatory patterns.

    \item \textbf{Mitigation strategies:} Upon identifying biases, developers should implement mitigation strategies \cite{pagano2023bias}. This can involve retraining the model with balanced datasets, fine-tuning specific parameters to reduce bias, or incorporating fairness constraints during the training process.

    \item \textbf{Adversarial training:} Adversarial training is a technique where the model is trained against adversarial examples designed to expose and correct biases \cite{xu2021robust}. This approach can enhance the model's robustness and fairness by forcing it to learn more nuanced and unbiased representations.

    \item \textbf{Data augmentation:} By augmenting the training data with synthetic examples that challenge existing biases, developers can help the model learn more equitable patterns and reduce bias in its outputs \cite{pastaltzidis2022data}.
\end{itemize}

These strategies collectively contribute to improving the fairness and robustness of ChatGPT, ensuring that its outputs are more equitable and less susceptible to biases that may exist in the training data. However, ongoing monitoring and refinement are essential to address emerging biases and maintain fairness as societal norms evolve.
\subsection{Privacy concerns}
\begin{figure}[h]
\centering
\tcbox[colframe=gray!30!black,
           colback=gray!20]{\includegraphics[width=0.77\textwidth]{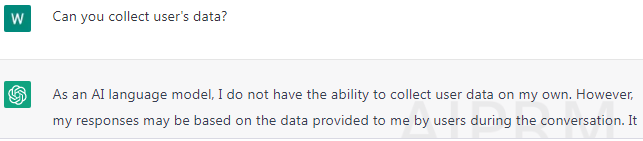}}
\caption{ChatGPT answers a privacy question.}
\label{fig:privacy}
\end{figure}

Privacy concerns are a potential issue when using ChatGPT. Although ChatGPT says that it doesn't collect user data as shown in Figure \ref{fig:privacy}, it may collect some information, such as chat logs, which could be used for malicious purposes or shared without user consent. Developers must ensure that user data is protected and handled responsibly in accordance with relevant privacy regulations. Users should also be aware of the potential privacy risks associated with using ChatGPT and take steps to protect their personal information.

Besides, as with any rapidly advancing technology, it is crucial to examine the possible ethical and societal consequences of ChatGPT and other large language models. Matters like privacy and employment impacts are just a few of the issues that must be thoroughly assessed as these technologies continue to progress. For instance, the use of large language models in customer service might lead to job loss in that industry, and the collection of data through these models raises serious concerns about privacy \cite{lund2023information}. Therefore, it is vital that we thoughtfully consider the ethical implications of these technologies and ensure that they are developed and utilized in a responsible and ethical manner \cite{aljanabi2023chatgpt}.
\subsection{Responsibility in deploying the tool}
Responsibility in deploying the tool refers to the ethical considerations that must be taken into account when using ChatGPT \cite{oviedo2023risks}. Developers must consider the potential impacts of their ChatGPT implementation, including the risks of unintended consequences, biases, and misuse. They must ensure that the tool is used in a responsible and ethical manner and take steps to minimize potential harm \cite{zhuo2023exploring}. This includes being transparent about the use of ChatGPT, obtaining user consent when appropriate, and establishing clear guidelines for its use. Additionally, developers should consider the potential social and ethical implications of their ChatGPT implementation, and take steps to mitigate any negative impacts. Ultimately, the responsible deployment of ChatGPT requires a comprehensive understanding of the tool's capabilities and limitations, as well as a commitment to ethical and responsible use.

\section{Prompt engineering and RAG}
\label{sec:prompt}
Prompt engineering and Retrieval-Augmented Generation (RAG) are powerful techniques that enhance the performance of large language models. This section explores how they work individually and together to produce more accurate, context-aware outputs.
\subsection{Prompts}
Prompts are a set of instructions that are provided to a large language model to ensure specific characteristics of the generated output, automate processes, and enforce rules \cite{zuccon2023dr}. They can be considered as a form of programming that can customize the interactions with the large language model and modify the outputs according to the desired requirements. Accordingly, prompt patterns are akin to software patterns as they provide repeatable solutions to specific issues, but they are tailored to the context of generating output from large-scale language models like ChatGPT. While software patterns offer a structured method for addressing common software development problems, prompt patterns provide a structured method for tailoring the output and interactions of language models. Therefore, to get the desired answers to our requests, prompt patterns should be improved \cite{white2023prompt,polak2024}. To do so, many solutions can be found to enhance the quality of the input and output of large language models. These improvements include question refinement, alternative approaches, cognitive verifier, and refusal breaker. The question refinement pattern ensures that the large language model always suggests an improved version of the user's question. The alternative approaches pattern requires the large language model to propose alternative ways of accomplishing a specific task specified by the user. The cognitive verifier pattern instructs the large language model to automatically suggest a set of sub-questions for the user to answer, which can then be combined to produce an answer to the main question. The Refusal breaker pattern requires the large language model to automatically rephrase the user's question if it cannot produce a response. Furthermore, context control is centered on managing the contextual information under which the large language model functions. This category comprises the context manager pattern that enables the user to define the context for the output generated by the large language model.

\begin{figure}[h]
\centering
\tcbox[colframe=gray!30!black,
           colback=gray!20]{\includegraphics[width=0.55\textwidth]{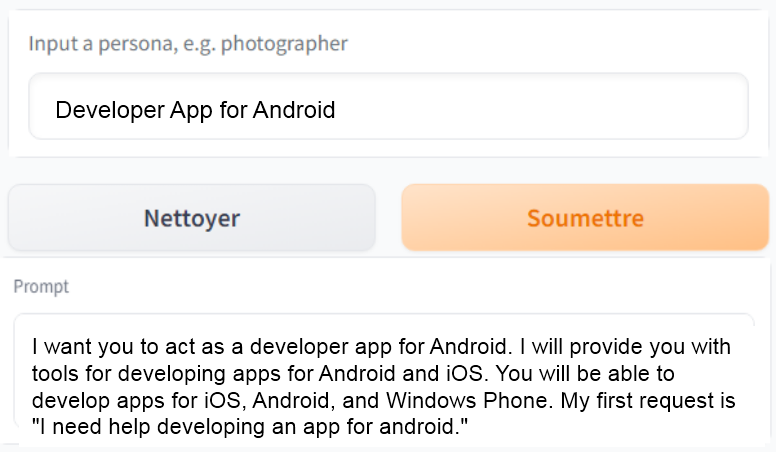}}
\caption{Example of prompt generation using ChatGPT-prompt-generator of Hugging face.}
\label{fig:prompt_gen}
\end{figure}

Additionally, to generate efficient prompts, Hugging face \cite{hugging1} introduced a new generator called "ChatGPT-prompt-generator", it is based on a BART model pre-trained on a prompt dataset called "Awesome-chatgpt-prompts" \cite{Prompt-Generator}.
To generate efficient prompts, the user should give an indication, and the generator will provide a convenient prompt for ChatGPT. Figure \ref{fig:prompt_gen} presents an example of prompt generation using Hugging face.

\subsection{Retrieval-Augmented Generation (RAG)} is a technique in large language models (LLMs) that enhances their ability to provide accurate and up-to-date information by combining document retrieval with text generation \cite{lewis2020retrieval}. Instead of relying solely on pre-trained knowledge, RAG retrieves relevant documents from an external knowledge base based on a user's query and then uses an LLM to generate a response grounded in those documents. This approach reduces hallucinations, improves factual accuracy, and allows the model to access dynamic or domain-specific information without retraining. RAG is particularly useful for applications like customer support \cite{xu2024retrieval}, medicine \cite{xiong2024benchmarking}, text generation \cite{li2022survey}, legal research \cite{wiratunga2024cbr,hindi2025enhancing}, and enterprise search, where precise and current information is critical. Recent improvements on RAG have been introduced, such as Graph RAG \cite{peng2024graph} and Active RAG \cite{jiang2023active}, which aim to enhance the retrieval and reasoning capabilities of traditional RAG systems. Graph RAG incorporates a structured graph of retrieved documents, enabling the model to understand relationships between different pieces of information and reason over interconnected content. Active RAG, on the other hand, introduces an iterative mechanism that actively evaluates the usefulness of retrieved documents and selectively refines queries to improve the relevance of information passed to the generator. These advancements push RAG beyond static retrieval, making it more adaptive, context-aware, and capable of handling complex information needs.

Agentic RAG (Agent-based Retrieval-Augmented Generation) is an advanced form of RAG that integrates autonomous agent behavior into the retrieval and generation process. Unlike standard RAG, where retrieval and generation are typically single-step and static, Agentic RAG allows the system to reason, plan, and take multiple actions over time to fulfill complex tasks. Table \ref{tab:rag_comparison} presents a comparison between standard RAG and agentic RAG.

\begin{table}[h!]
\centering
\caption{Comparison between Classical RAG and Agentic RAG.}
\begin{tabular}{|p{4cm}|p{5cm}|p{5cm}|}
\hline
\textbf{Aspect} & \textbf{Standard RAG} & \textbf{Agentic RAG} \\
\hline
\textbf{Process type} & Single-step retrieval and generation & Multi-step, iterative reasoning and retrieval \\
\hline
\textbf{Control flow} & Fixed pipeline & Dynamic, agent-driven with planning and decision-making \\
\hline
\textbf{Retrieval strategy} & One-time retrieval before generation & Adaptive retrieval with feedback and refinement \\
\hline
\textbf{Capabilities} & Static question answering & Complex task solving, tool use, and reasoning \\
\hline
\textbf{Memory and state} & Stateless, no memory of past steps & Maintains memory/state across actions \\
\hline
\textbf{Tool integration} & Limited to retriever and generator & Flexible tool use (retrievers, APIs, calculators, etc.) \\
\hline
\textbf{Use cases} & Simple Q\&A, document search & Multi-step tasks, research agents, autonomous assistants \\
\hline
\end{tabular}

\label{tab:rag_comparison}
\end{table}

\begin{figure}[h]
\centering
\tcbox[colframe=gray!30!black,
           colback=gray!20]
{\includegraphics[width=0.90\textwidth]{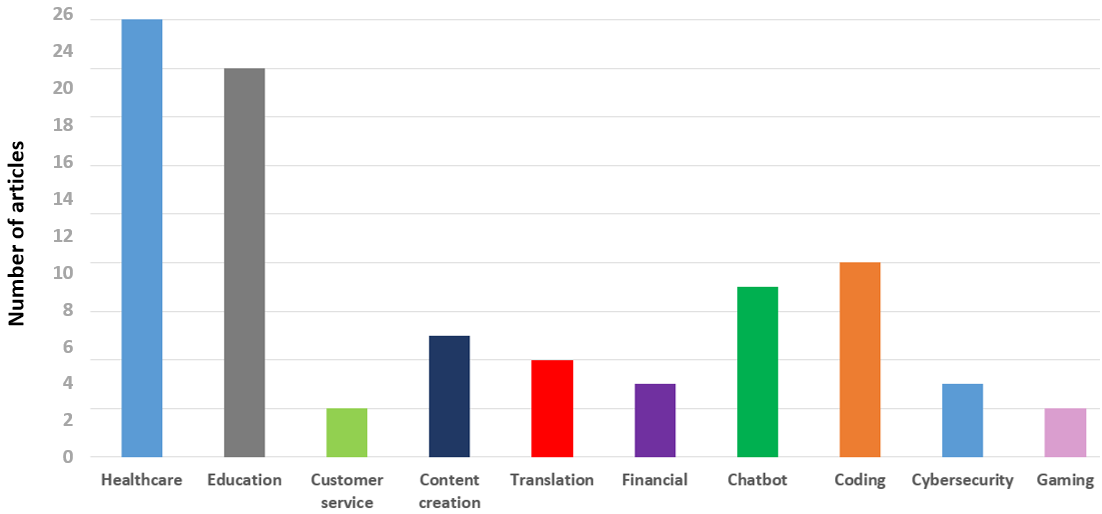}}
\caption{Overview of the categories covered in the reviewed papers.}
\label{fig:domains}
\end{figure}

\section{Future directions for ChatGPT and natural language processing}
\label{sec:future1}
In this review study, we highlight the main domains that have been quickly affected by the emergence of ChatGPT. Figure \ref{fig:domains} presents an overview of the categories reviewed in this paper where healthcare and education have been the most cited.
Future directions for ChatGPT and natural language processing involve ongoing research and development to improve the tool's capabilities and overcome its limitations. Some potential areas of focus include improving the ability of ChatGPT to understand the context and generate more natural language, developing more effective techniques for training and fine-tuning the tool, and addressing issues related to bias and fairness. Other potential directions for natural language processing include developing more advanced dialogue systems, enhancing the ability of ChatGPT to handle multiple languages and dialects, and integrating the tool with other AI technologies such as computer vision and speech recognition \cite{qin2023chatgpt,kumari2024chatgpt}. As the field of natural language processing continues to evolve, there is significant potential for ChatGPT to become an even more powerful tool for understanding and interacting with human language.

\section{Future directions for ChatGPT in vision domain}
\label{sec:future2}
While ChatGPT is primarily a tool for natural language processing, there is potential for it to be integrated with computer vision technologies in the future. This could enable ChatGPT to perform more advanced tasks that involve both language and visual information, such as image captioning, visual question answering, and scene understanding. For example, ChatGPT could be used to generate descriptions of visual content based on user input, or to answer questions about images and videos.
One potential direction for ChatGPT in the vision domain is the development of more advanced multimodal models that can integrate both language and visual information. This could involve the use of techniques such as attention mechanisms and graph neural networks to enable ChatGPT to better understand the relationships between visual elements and language concepts. In \cite{maddigan2023chat2vis}, a natural language interface solution is proposed that suggests a different approach from the conventional method of developing new versions of language models. Instead of training models from scratch, it advocates leveraging the advancements in pre-trained large language models such as ChatGPT and GPT-3 to convert natural language into executable code, which can then be used to generate appropriate visualizations. The study systematically evaluates the effectiveness of GPT-3, Codex, and ChatGPT across multiple case studies, assessing their ability to interpret user intent and produce accurate and meaningful visual representations. Furthermore, the performance of these models is benchmarked against previous research in visualization, highlighting their strengths and limitations in handling complex queries, generating diverse visualization types, and adapting to domain-specific requirements.

Another potential area of focus is the development of more advanced transfer learning techniques that enable ChatGPT to leverage knowledge learned from both language and visual domains. This would enhance its ability to process and generate responses that integrate textual and visual information, making it more effective for multimodal applications. By refining these techniques, ChatGPT could improve its performance in tasks such as image captioning, video analysis, and cross-modal reasoning. This could involve pretraining ChatGPT on large-scale datasets that combine both text and image data, such as the Conceptual Captions dataset. By doing so, ChatGPT could learn to understand the relationships between language and visual information in a more comprehensive way, enabling it to perform more complex tasks in the vision domain. Thus, the integration of ChatGPT with computer vision technologies represents an exciting area of future research that could lead to new applications and capabilities in the field of AI such as artistic creations tasks like painting \cite{guo2023can}, intelligent vehicle and driving \cite{du2023chat,gao2023chat,zhang2023hivegpt}, industry \cite{wang2023chat}, conversational applications with visual human-machine interaction \cite{wang2023chatgpt2}.

\section{Conclusion} 
ChatGPT is a powerful tool in the field of artificial intelligence that has significant implications for society. This review compares ChatGPT to other LLMs like LLaMA 3, Gemini, and Deepseek, highlighting their unique strengths, architectures, and performance in various tasks. While LLaMA 3 is known for its efficiency and open-source accessibility, Gemini excels in integrating multimodal capabilities, and Deepseek focuses on domain-specific applications. ChatGPT’s ability to generate human-like language has already led to numerous real-world applications, such as chatbots and language translation tools. However, as with any technology, there are potential ethical concerns that must be addressed in order to ensure that its impact on society is positive. One of the most important ethical considerations is the potential for bias and harmful language patterns to be perpetuated by ChatGPT. It is critical that developers and users of the tool take steps to mitigate these risks and ensure that ChatGPT is used in an ethical and responsible manner. This includes addressing issues related to fairness and bias in training data, as well as taking steps to protect user privacy and ensure that the tool is used in a way that aligns with social values. At the same time, there is significant potential for ChatGPT to have a positive impact on society. By enabling more natural and intuitive interactions with technology, ChatGPT has the potential to enhance communication and collaboration across a wide range of domains, from healthcare and education to business and entertainment. As the tool continues to evolve and improve, there is no doubt that it will play an increasingly important role in shaping the future of human-computer interaction.

Finally, the effectiveness of ChatGPT largely depends on the quality of input prompts. Clear and specific prompts are critical in guiding the system to generate accurate and relevant outputs, while also addressing potential limitations and biases. Proper use of prompts is essential for maximizing the potential of ChatGPT in NLP applications. Additionally, continuous refinement of prompt engineering techniques can further enhance the model’s performance across diverse contexts. As users become more adept at crafting effective prompts, the reliability and utility of ChatGPT in real-world scenarios will continue to improve.
\bibliographystyle{unsrt}  
\bibliography{references}

\end{document}